%% file: acl_latex.tex
\pdfoutput=1

\documentclass[11pt]{article}

\usepackage{acl}

\usepackage{times}
\usepackage{latexsym}
\usepackage{amsmath}
\usepackage{multirow} 
\usepackage[table]{xcolor}
\usepackage[T1]{fontenc}
\usepackage{makecell}

\usepackage[utf8]{inputenc}
\usepackage{listings}
\usepackage{enumitem}
\usepackage{microtype}

\usepackage{inconsolata}

\usepackage{graphicx}
\usepackage{changepage}
\usepackage{booktabs}

\usepackage{seqsplit}

\title{Multilingual Prompting for Improving LLM Generation Diversity 
}

\author{
 \textbf{Qihan Wang\textsuperscript{1}},
 \textbf{Shidong Pan\textsuperscript{1,2}},
 \textbf{Tal Linzen\textsuperscript{1}},
 \textbf{Emily Black\textsuperscript{1}}
\\
\\
 \textsuperscript{1}New York University,
 \textsuperscript{2}Columbia University
\\
  \texttt{\{qw2488, shidong.pan, linzen, emilyblack\}@nyu.edu}
}

\begin{document}
\maketitle

\begin{abstract}
Large Language Models (LLMs) are known to lack cultural representation and overall diversity in their generations, from expressing opinions to answering factual questions.
To mitigate this problem, we propose \emph{multilingual prompting}: a prompting method which generates several variations of a base prompt with added cultural and linguistic cues from several cultures, generates responses, and then combines the results.
Building on evidence that LLMs have language-specific knowledge, multilingual prompting seeks to increase diversity by activating a broader range of cultural knowledge embedded in model training data. Through experiments across multiple models (GPT-4o, GPT-4o-mini, LLaMA 70B, and LLaMA 8B), we show that multilingual prompting consistently outperforms existing diversity-enhancing techniques such as high-temperature sampling, step-by-step recall, and persona prompting. 
Further analyses show that the benefits of multilingual prompting vary between high and low resource languages and across model sizes, and that aligning the prompting language with cultural cues reduces hallucination about culturally-specific information. 
\end{abstract}

\section{Introduction}
\input{emilys_version/introduction}

\section{Related Work}
\input{emilys_version/related_work}
\section{Multilingual and Multicultural Prompting}
\input{emilys_version/multilingual}

\section{Increasing Demographic and Perspective Diversity}
\label{sec:diversity}
\input{emilys_version/experiment_v2}
\subsection{Results}
\input{emilys_version/result_diversity}

\section{Language Helps Prevent Hallucination}
\input{emilys_version/hallucination}

\section{Multilingual Prompting Across Resource Levels}
\label{sec:hallucination}
\input{emilys_version/high_low}

\section{Conclusion}
\input{conclusion}

\bibliography{reference}

\appendix

\section{Appendix}
\input{appendix}

\end{document}

%% file: emilys_version/introduction.tex
\begin{figure}[t]
    \centering
    \includegraphics[width=1\linewidth]{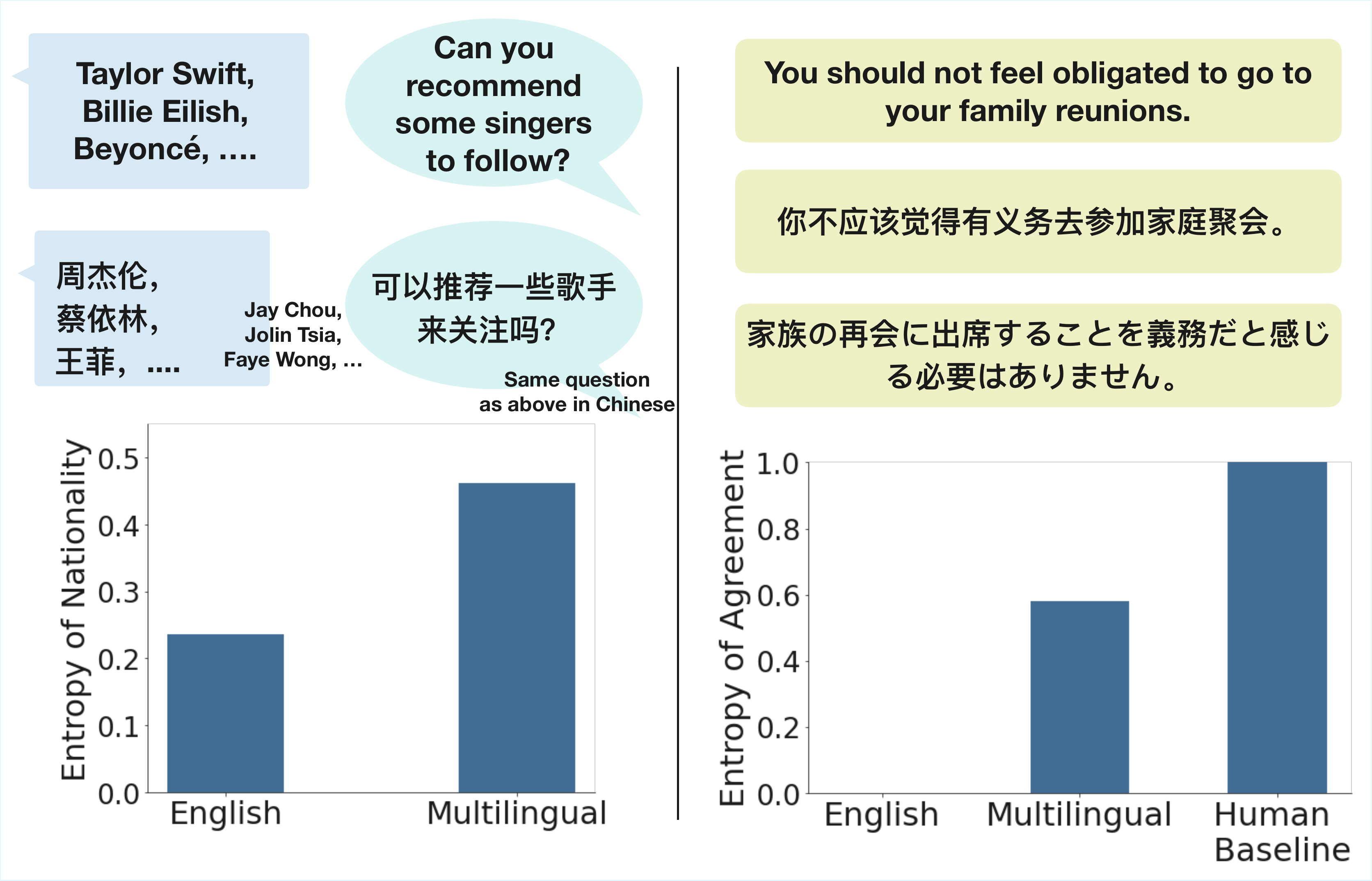}
    \caption{An example of the diversity of an LLM's (GPT-4o) responses when prompted in English versus in multiple languages: on the left, we show demographic diversity, specifically the range of different nationalities represented in an answer about which singers to follow; on the right, we show the level of agreement with a controversial social norms statement. We measure diversity by calculating the (normalized) entropy of model responses, explained in more detail in Section~\ref{sec:metrics}. Multilingual prompting leads to an increase in diversity.} 
    \label{fig:example}
\end{figure}

\begin{figure*}[t]
\centering
\includegraphics[width=\textwidth]{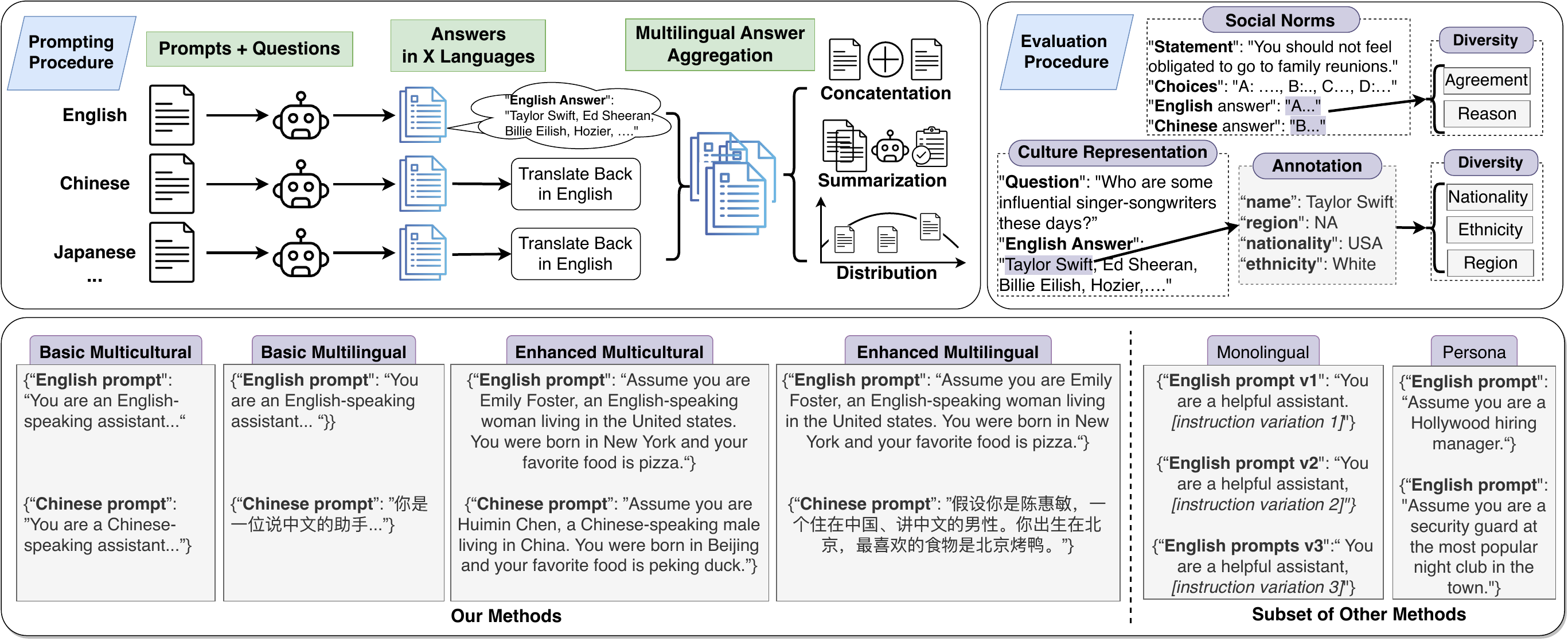}
\vspace{-15pt}
\caption{Above: an overview of multilingual and multicultural prompting, and our diversity evaluation. Below: example prompts from our multilingual and multicultural methods, and a subset of methods we compare to.}
\label{fig:process}
\vspace{-10pt}
\end{figure*}

Large Language Models (LLMs) are now omnipresent: they have effectively replaced traditional search engines, and people use them to do everything from studying to planning travel and other leisure activities~\cite{chatterji2025people}. As a result, LLMs have an ever-increasing power to dictate exposure to ideas, facts, and people, as the public uses LLMs to gain access to information. It is important that this exposure is distributed in an equitable manner. Lack of diversity in LLM generations---especially when querying for new information---can lead to a host of problems: lack of demographic diversity when the LLM is queried about individuals can lead to unfair lack of exposure of artists, academics, and other professionals on the basis of their race, ethnicity, or nationality. Lack of cultural diversity in response to questions about controversial topics can contribute to inaccurate results when LLMs are used as substitutes for human responses in user studies, annotation tasks, and opinion surveys, as these responses do not reflect the diversity of real-world perspectives.
Indeed, prior work has shown that LLMs do not represent the true diversity of human expression in a variety of ways---from reducing sentiment and topic diversity for tasks such as book reviews~\cite{wu2024generative}, to demonstrating poor linguistic diversity when helping humans write essays~\cite{padmakumar2023does}. Perhaps even more importantly, LLMs have been shown to generate largely monocultural responses to controversial questions, often leaning toward expressing Western values~\cite{wang2025large}, or even a subset of Western values~\cite{santurkar2023whose}. 

These trends continue in our own experiments. In Figure~\ref{fig:example}, we show an example of LLM responses when asked about individuals in various professions, for example, what musical artists to listen to. When prompted in English, LLM answers are largely limited to American artists, and exclude those from other cultural backgrounds.
Similarly, when we ask LLMs in English whether they agree with a 
statement known to be controversial among humans~\cite{forbes2020social}, e.g., “You should not feel obligated to go to your family reunions”--- the models largely agree with this statement, generating homogeneous responses which do 
not reflect the variety of perspectives across different cultural contexts. 

In this work, we propose that language and other cultural cues can be a powerful lever for enhancing diversity in LLM outputs, which points to a way to mitigate these problems. 
Returning to Figure~\ref{fig:example}, we see that prompting the model separately in multiple different languages and combining the responses leads to higher diversity in the ethnicity and nationality of the artists suggested.  Similarly, if we ask the model in several different languages for opinions about whether people should feel obligated to attend their family reunions, the response varies much more in its level of agreement with the statement. 
These results add to increasing evidence~\cite{aggarwal2025language,hamalainen2023evaluating} that LLMs encode culturally specific information linked to the language and other cultural cues in the input--—and we suggest
these differences in LLM behavior across different languages and cultural cues present an opportunity to deliberately create more diverse generations. 

But this raises the question: what is the best way to prompt the model to tap into its culture-specific knowledge, in order to create more diverse, but correct, generations? 
Is language itself the best signal to prompt the model to dip into particular cultural knowledge, or are cultural cues such as giving a name, birthplace and personality cues for a persona on their own enough? (See Figure~\ref{fig:process} for example prompts.) In Sections~\ref{sec:diversity} and~\ref{sec:hallucination}, we explore these questions. We find that both language and cultural cues are important for boosting diversity, but prompting in the language connected to a given culture achieves higher diversity overall. Further, we see that matching cultural cues and language is important to prevent hallucination for culturally relevant information, e.g. giving the names of actual Chinese singers as opposed to simply outputting Chinese names.

Given these results, we posit that \emph{multilingual} prompting, using cultural cues and language, is a preferable method to \emph{multicultural} prompting, which uses cultural cues alone while prompting in English. After establishing this result, in Section~\ref{sec:high_low}, we investigate how multilingual prompting performs as the number of languages increases, as well as over low- and high-resource languages.

In sum, in this work, we present the following three contributions: (1) we introduce and evaluate \emph{multilingual} and \emph{multicultural prompting} as shown in Figure~\ref{fig:process} as methods to increase demographic, cultural, and other forms of diversity in LLM generations. We find that these methods increase demographic and cultural diversity in LLM generations better than state of the art methods, such as step-by-step recall prompting~\cite{hayati2023far}, generating personas~\cite{wang2025large}, and increasing temperature~\cite{chung2023increasing}, all while maintaining accuracy on factual tasks. (2) We explore whether 
using the native language that corresponds to the cultural cues reduces hallucination for culture-specific pieces of information, such as names of famous singers from different parts of the world.
Based on human evaluation of model outputs, we find that specifically prompting in the language associated with a specific culture reduces hallucinations about that culture when compared to prompting in English, suggesting that language is imperative for generating \emph{accurate} and diverse information.
(3) Finally, we evaluate the performance of multilingual prompting as the number of languages increases, as well as across lower- and high-resourced languages. 
We see that overall, the diversity gain from multilingual prompting increases with the number of languages used.
Further, we see that some models gain more diversity from prompting in high-resourced languages, while smaller models demonstrate greater diversity gains from lower-resourced languages.


%% file: emilys_version/related_work.tex
\noindent\textbf{Current Diversity Issues in LLMs.}
Recent research has raised concerns about the lack of diversity in LLM opinions, cultural perspectives, and linguistic expression~\cite{wang2025large,padmakumar2023does,Tevet2020EvaluatingTE}.
For example, recent work has revealed that LLMs reflect the opinions of dominant groups disproportionately 
even despite prompt steering \citep{santurkar2023whose}, and that LLMs can produce nearly identical responses even when primed with demographic variation in prompts~\citep{park2024diminished, Kitadai2024ExaminingTF}. More broadly, many authors have expressed concern about homogenizing, monocultural tendencies of LLMs leading to societal harm, from discrimination to model collapse~\citep{bommasani2022picking, fabris2022algorithmic, Kleinberg2021AlgorithmicMA, wu2024generative,shumailov2024ai, zhang2024forgotten}.

To counter these issues, researchers have explored methods to increase diversity in LLM outputs while maintaining coherence and accuracy. We compare multilingual prompting to many of these methods in Section~\ref{sec:result}, including sampling-based approaches (e.g., high temperature, top-$k$ sampling); persona-based prompting \citep{cheng2023marked}, where models simulate varied viewpoints by adopting socio-demographic roles or synthetic identities \citep{mukherjee2024cultural, beck2023sensitivity}; and step-by-step recall prompting, which encourages the model to explore multiple evaluative dimensions or iteratively expand its answer space \citep{hayati2023far}. Overall, based on our evaluation of demographic diversity for prompts about individuals and diversity of perspective in prompts on social norms, we find that multilingual prompting is more effective than these other methods. 

\vspace{0.5em}
\noindent\textbf{LLMs Across Languages.} 
A separate line of work has shown that LLMs perform variably across languages~\cite{ohmer2023separating, goldman2025eclektic}. While much of this work has focused on negatives—--e.g., 
showing that LLMs have differing ability to recall facts in different languages---we argue that this variability can be 
exploited. Perhaps most related to our work,
Kwok et al. explore to what extent language and other cultural cues can help LLMs respond to questions in a manner that reflect a \emph{particular} cultural background~\cite{kwok2024evaluating}.  Importantly, our work differs in that we suggest multilingual prompting as a method to improve general diversity in LLM responses, rather than attempting to faithfully recreate a particular cultural background. Interestingly, their findings suggest that using native language is not helpful for eliciting representative responses for specific cultures, but 
that culture-and nationality-specific cues in English are most effective. However, 
we find 
that adding native language provides a diversity boost when used in conjunction with cultural cues. Further, while \newcite{kwok2024evaluating} find that using native language decreases performance of matching human outputs from a given culture, we find that using native language \emph{increases} the performance of the LLM by decreasing culture-specific hallucination (see Section~\ref{hallucination}). 

%% file: emilys_version/multilingual.tex

We present two related prompting methods in this work, which we call multilingual and multicultural prompting. Both multilingual and multicultural prompting work to increase LLM generation diversity by eliciting responses to several different versions of the same prompt, each with different cultural and/or linguistic cues, and then combining them into one response.
One goal of this work is to understand which method is the best to increase diversity in LLM generations.
\emph{Multicultural} prompting does so by relying solely on adding cultural cues, in English---such as adding to the prompt that the LLM is English-speaking, or giving a persona a Chinese name and adding they were born in Beijing. For \emph{multilingual} prompting, we rely on these cultural cues \emph{and} translate the prompt to the language associated with that culture. See Figure~\ref{fig:process} for examples.
Our multilingual and multicultural prompting methods consist of three main steps, also shown in Figure~\ref{fig:process}:

\paragraph{1. Preparing the Queries:} We begin by editing the original English query by creating $n$ versions of the original query, each with added cues related to various languages or culture (e.g., ``You are a Chinese-speaking assistant'', see more in Figure~\ref{fig:process}) and, in the case of multilingual prompting, also translating the prompt into the corresponding target languages (e.g. Chinese). For example, to do multilingual or multicultural prompting with English, Chinese, and Japanese, we generate three versions of the prompt, each corresponding to one language and set of cultural cues.

We have two types of multicultural/lingual queries, one set of which we label ``basic'' and the other we label ``enhanced''. 
The basic variant consists of prompting the model with ``You are an [language]-speaking assistant''. Enhanced multicultural/lingual prompting adds three addition cultural cues: a name, birthplace, and favorite food.
Following prior work~\cite{kwok2024evaluating}, in preliminary experiments, we find that language completely on its own, without any cultural signal, does not increase diversity.

In the implementation we release,\footnote{Our code repository is available at:  https://github.com/mangocyann/Multilingual-Prompting-for-Improving-LLM-Generation-Diversity.} users can select arbitrary target languages to suit their own cultural preferences. The experiments presented in this paper focus on Chinese, Japanese, and English in our initial experiments in Sections \ref{sec:main_exp}, before expanding to Spanish, French, Nepali, Thai, Turkish, and Ukrainian in section \ref{sec:high_low}. 
    
\paragraph{2. Model Response Generation:} The modified prompts (one per language) are then given to the LLM one at a time. The model generates responses for each modified query.
    For multilingual prompting, the model responds in various different languages, and we translate all answers back into English using GPT-4o-mini. 
    
\paragraph{3. Aggregation:} 
We then combine the responses into one answer. In this work, we simply concatenate the responses from the modified prompts to generate one overall response. We make this choice easily by tabulating diversity by comparing the range of responses from multilingual prompting to other prompting methods, such as our baseline of rephrasings of the original query in English.
We discuss other methods for aggregation of LLM responses in our discussion section. 

%% file: emilys_version/experiment_v2.tex
\label{sec:main_exp}
In this section, we present our experimental framework and results showing how multilingual prompting can increase demographic and perspective diversity compared to other 
methods. 

\subsection{Dataset and Metrics}
\label{sec:experiment_setup}

We test the diversity of LLM responses for two tasks: responding to questions about controversial social norms, and naming individuals in various professions. 

\vspace{0.5em}
\noindent\textbf{Social Norms: the Social Chemistry
101 Dataset.}
For this task, we ask the LLM to select a multiple-choice response to a statement making a claim about a controversial social norm---i.e., one that humans have differing opinions on. 

Specifically, we use a modified version of  
the Social Chemistry 101 dataset \cite{forbes2020social}, a large-scale corpus designed to capture human moral and social norms through Rules of Thumb---natural language statements reflecting implicit principles of social acceptability.
The dataset includes human baseline judgments, where annotators indicate the degree of agreement with each statement.
In order to ensure we are focusing on truly controversial statements where diverse responses are desirable, we specifically focus on a subset of the dataset where only 25--75\% of annotators agree with the given statement. 
Example statements include ``\textit{You should not feel obligated to go to your family reunions}'' and ``\textit{It's wrong to keep your health a secret}''. 

While the original dataset only includes the statements themselves, in order to evaluate diversity in model agreement and model reasoning, we used an LLM to generate multiple-choice responses to each statement. 
Specifically, we generated four choices, 
including two agreeing and two disagreeing valences, all with explanations that encoded different moral values to support the viewpoints.
For example, \textit{``valence'': ``agree''; ``explanation'': ``Obligations shouldn't dictate personal choices; you should only go if you want to.'' }
More examples and the prompts used to generate them, are in Appendix~\ref{generating_choices}. 
We then use these multiple-choice answers to calculate the diversity in both the judgment (agreeing or disagreeing) and in the justification for it.



\vspace{0.5em}
\noindent\textbf{Demographic Representation: the People Diversity Dataset.} 
In this task, we ask an LLM to generate names for individuals in various professions, and assess the demographic diversity of the response 
in terms of nationality, ethnicity, and geographic region. 
To do so, we use the People Diversity Dataset~\cite{lahoti2023improving}.
The dataset consists of 105 questions asking for the names of individuals 
in a particular profession and context.
For example, ``\textit{Name some writers that inspire you}''. 
In our experiment, we ask the LLMs to generate exactly 10 names for each question.


\paragraph{Metrics.} 
\label{sec:metrics}
To assess the diversity of LLM responses to questions from both datasets, we rely on the following entropy measures.


\noindent\textit{Reason and Valence Entropy.} To assess the diversity in reasoning across LLM responses to social norms questions, for each prompting strategy, we calculate the average entropy across the three responses we generate from the model for each query corresponding to each language or culture. We call this \emph{reason entropy}. 
To assess the diversity in agreement/disagreement, we calculate entropy, but treat responses that have the same judgment
(agree/disagree) as interchangeable. We call this \emph{valence entropy}. For example, Reason Entropy is calculated as
$H_{\text{Reason}} = - \sum_{i \in \{A,B,C,D\}} p(i) \log p(i)$, where \( p(i) \) represents the probability of the model selecting choice \( i \). Valence entropy only has two choices: agree or disagree. A higher entropy indicates a greater diversity, because we focus on controversial questions, higher diversity is generally desirable.

\paragraph{Demographic Entropies.}
To evaluate demographic representation, we use an LLM (GPT-4o-mini) to annotate the nationality, ethnicity, and region for each name generated. To ensure the reliability of these cultural origin annotations, we conduct performance checks for the annotation (see details in Appendix~\ref{apppendix:annotation_demographic}). Then, for each question, we calculate the entropy for each attribute across the thirty names (10 names for each of the 3 languages) generated from each prompting strategy to measure the cultural diversity of the model’s predictions.
For each attribute \( A \), we define its entropy \( H(A) \) as: 
$
H(A) = - \sum_{c \in C_A} p(c) \log p(c)$, 
where \( C_A \) is the set of possible categories within attribute \( A \), and \( p(c) \) represents the probability of category \( c \) occurring in the model's annotations. In Section~\ref{sec:result}, we report the average normalized entropy across all questions.


To place all metrics on a common \([0,1]\) scale, we divide each raw score by the \emph{maximum} value it could theoretically attain under the same option count,
$\widetilde{H} \;=\; \frac{H}{H_{\max}}$,
where \(H\) is the unnormalized value and \(H_{\max}\) is the corresponding upper bound. 
Details about normalization 
can be found in Appendix~\ref{appendix:normalization} and Appendix~\ref{name_normalization}.

\subsection{Baselines and Performance Tests}
To ensure a fair comparison across prompting strategies, we generate three LLM responses with each strategy (multicultural, multilingual, and the baselines and comparison methods below), and evaluate the diversity of the concatenated responses.
The exact phrasing of all prompts is included in Appendix~\ref{prompt_socialnorm} and Appendix~\ref{prompt_culture}.


\noindent\textit{Baseline.}
Our baseline consists of prompting the model $n$ times in English only. 
To ensure comparability with multilingual methods, we adopt the same sampling strategy: each query is preceded by a short preamble (``You are a helpful assistant'') and rephrased into multiple lexical variants with high syntactic similarity, as prior work has shown that LLMs are sensitive to phrasing~\cite{sclar2023quantifying}. 
Then, the $n$ outputs are concatenated and used to compute diversity metrics. 
We refer to this as \emph{monolingual prompting}.
By comparing such monolingual and multilingual prompting, we can attribute observed entropy gains to cross-lingual variation rather than the phrasing sensitivity of LLMs in generation. (See Figure~\ref{fig:process} and Appendix~\ref{prompt_socialnorm} and~\ref{prompt_culture} for prompt details.)

\paragraph{Comparisons.} 
To assess the effectiveness of our approach, we compare our method against previously established diversity-enhancing techniques:
\begin{itemize} [leftmargin=*, noitemsep]
    \item High-temperature sampling, using the monolingual strategy from above, but setting temperature = 1.3 \cite{chung2023increasing}. 
    \item Requesting Diversity: We also compare with prompts that simply ask the model to be diverse, namely by adding ``Please try to be as diverse as possible'' to the monolingual prompt. 
    For these two comparison methods, to increase diversity, we also evaluate the diversity over concatenated responses of three rephrased versions of the prompt.
    \item Random Personas: Following prior work \cite{wang2025large}, we create personas for the model prior to prompting. To separate persona prompting from multilingual prompting, these prompts do not encode cultural information, but rather professions and other personality traits. We use the same number of personas as languages and evaluate concatenated responses. 
    \item Step-by-step Recall~\cite{hayati2023far}: This prepends past answers to subsequent questions sequentially to ask the model to generate new answers after reflection on prior answers. To compare fairly, we generate query responses from three rounds of Step-by-Step Recall, and evaluate the concatenated responses.
\end{itemize}

We include Step‑by‑Step Recall and Requesting Diversity for the demographic diversity tasks, but not social norm tasks, as they do not work well with multiple-choice outputs. Step‑by‑Step Recall asks the model to reveal its first answer and then generate a different one in the next round, forcing the model to change its mind, which contradicts the spirit of a single‑choice multiple‑choice task. Similarly,  Requesting Diversity is designed to elicit a set of varied outputs, but in the social‑norm setting, the model must commit to exactly one label, so the notion of “being diverse” reduces to a single token and loses its intended effect. 

\paragraph{Performance Checks.} To ensure that the LLMs are reasoning well 
when responding to the multiple-choice questions given from the modified social norms dataset~\cite{forbes2020social}, we perform a test with different multiple-choice responses based on Zellers et al.~\cite{zellers2019hellaswag}, where three out of four responses are logically nonsensical reasons for agreeing or disagreeing to the controversial statement, discussed further in Appendix~\ref{appendix:sanity_check}. More broadly, to verify that multilingual prompting does not compromise the factual accuracy of language models, we evaluate their performance on the Multilingual Grade School Math Benchmark {(MGSM)~\cite{shi2022language}}, which consists of mathematical reasoning tasks translated into multiple languages. Across all models, we observe that multilingual prompting maintains similar factual accuracy to monolingual prompting: GPT-4o-mini shows virtually no change; for GPT-4o and LLaMA-70B, there is a slight  performance drop around 5\%, but the overall competency of the model remains intact. More information is in {Appendix~\ref{sec:factuality}}. 

\paragraph{Models.} We conduct experiments over four mainstream models: \texttt{GPT-4o}, \texttt{GPT-4o-mini}~\cite{hurst2024gpt}, \texttt{LLaMA 3.3 70B} and \texttt{LLaMA 3.1 8B}~\citep{grattafiori2024llama}.

%% file: emilys_version/result_diversity.tex
\definecolor{lightpurple}{RGB}{230, 230, 250}

\begin{table}[t]

\centering

\resizebox{1.0\linewidth}{!}{%
\begin{tabular}{llccc}
\hline
\textbf{Model} & \textbf{Strategy} & \textbf{Reason} & \textbf{Agreement} & \textbf{Demo Avg.} \\
\hline
\multirow{11}{*}{GPT-4o}
 & Monolingual (Baseline) & 0.079 & 0.076 & 0.315 \\
 \rowcolor{gray!15} & \textbf{Diversity-Enhancing} & & & \\
 & Requesting Diversity & --- & --- & 0.370 \\
 & High Temperature & 0.161 & 0.128 & 0.344 \\
 & Step-By-Step Recall & --- & --- & 0.378 \\
 & Random Personas & 0.166 & 0.150 & 0.335 \\
 \rowcolor{gray!15} & \textbf{Our Prompting} & & & \\
 & Basic Multicultural & 0.191 & 0.172 & 0.360 \\
 & Basic Multilingual & 0.249\textsuperscript{*} & 0.210 & \textbf{0.415} \\
 & Enhanced Multicultural & 0.280\textsuperscript{*} & 0.245\textsuperscript{*} & 0.378 \\
 & Enhanced Multilingual & \textbf{0.300}\textsuperscript{*} & \textbf{0.247}\textsuperscript{*} & 0.387 \\
\hline
\multirow{11}{*}{\shortstack{GPT \\ 4o-mini}}
 & Monolingual (Baseline) & 0.089 & 0.050 & 0.314 \\
 \rowcolor{gray!15}&  \textbf{Diversity-Enhancing} & & & \\
 & Requesting Diversity & --- & --- & 0.349 \\
 & High Temperature & 0.121 & 0.058 & 0.345 \\
 & Step-By-Step Recall & --- & --- & 0.363 \\
 & Random Personas & 0.128 & 0.088 & 0.338 \\
 \rowcolor{gray!15} & \textbf{Our Prompting} & & & \\
 & Basic Multicultural & 0.127 & 0.096 & 0.402 \\
 & Basic Multilingual & 0.299\textsuperscript{*} & 0.176\textsuperscript{*} & \textbf{0.426}\textsuperscript{*} \\
 & Enhanced Multicultural & 0.167 & 0.102 & 0.390 \\
 & Enhanced Multilingual & \textbf{0.304}\textsuperscript{*} & \textbf{0.190}\textsuperscript{*} & 0.413\textsuperscript{*} \\
\hline
\multirow{11}{*}{\shortstack{LLaMA \\ 70B}}
 & Monolingual (Baseline) & 0.050 & 0.048 & 0.311 \\
 \rowcolor{gray!15} & \textbf{Diversity-Enhancing} & & & \\
 & Requesting Diversity & --- & --- & 0.341 \\
 & High Temperature & 0.068 & 0.056 & 0.357 \\
 & Step-By-Step Recall & --- & --- & 0.359 \\
 & Random Personas & 0.135 & 0.122 & 0.312 \\
 \rowcolor{gray!15}& \textbf{Our Prompting} & & & \\
 & Basic Multicultural & 0.105 & 0.086 & 0.377 \\
  & Basic Multilingual & 0.262\textsuperscript{*} & 0.218\textsuperscript{*} & 0.402\textsuperscript{*} \\
 & Enhanced Multicultural & 0.280\textsuperscript{*} & 0.170 & 0.409\textsuperscript{*} \\
 & Enhanced Multilingual & \textbf{0.304}\textsuperscript{*} & \textbf{0.222}\textsuperscript{*} & \textbf{0.428}\textsuperscript{*} \\
\hline
\multirow{11}{*}{\shortstack{LLaMA \\ 8B}}
 & Monolingual (Baseline) & 0.094 & 0.064 & 0.325 \\
 \rowcolor{gray!15} & \textbf{Diversity-Enhancing} & & & \\
 & Requesting Diversity & --- & --- & 0.322 \\
 & High Temperature & 0.236 & 0.164 & --- \\
 & Step-By-Step Recall & --- & --- & 0.377 \\
 & Random Personas & 0.143 & 0.086 & 0.334 \\
 \rowcolor{gray!15} & \textbf{Our Prompting} & & & \\
 & Basic Multicultural & 0.257 & 0.208 & 0.380 \\
  & Basic Multilingual & \textbf{0.555}\textsuperscript{*} & 0.465\textsuperscript{*} & \textbf{0.427}\textsuperscript{*} \\
 & Enhanced Multicultural & 0.164 & 0.070 & 0.382 \\
 & Enhanced Multilingual & 0.471\textsuperscript{*} & \textbf{0.469}\textsuperscript{*} & 0.388 \\
\bottomrule
\end{tabular}
}%
\vspace{-5pt}
\caption{Normalized entropy across social norm (Reason, Agreement) and demographic representation (Demo Avg.). ``Demo Avg.'' stands for the demographic average between nationality, ethnicity, and region. 
`---' indicates experiments not run, explained in Section~\ref{sec:experiment_setup}. 
{*} indicates the statistically significant differences between our methods and the best performance in diversity-enhancing comparisons.
\vspace{-10pt}}
\label{tab:merged_diversity_final}
\end{table}
\label{sec:result}

\paragraph{Multilingual Prompting Boosts Diversity of LLM Responses.} To evaluate whether and how multilingual and multicultural prompting promotes more opinion diversity across social norm-related questions, and demographic diversity in questions about individuals, 
we compare LLM responses across prompting strategies using the three metrics defined earlier: Reason Entropy, Agreement Entropy, and Demographic Entropies. Table \ref{tab:merged_diversity_final} reports the mean normalized entropy scores for each model across the different prompting strategies. Strategies are grouped into baseline, comparison, and multilingual and multicultural (our) methods. Due to space constraints, 
we present the average results for nationality, ethnicity, and geographic region diversity. Full results, including graphs of table results for ease of interpretation, are in Appendix~\ref{appendix:full_result}.  

Across all models and metrics, multilingual prompting strategies consistently yield the highest diversity scores. Enhanced multilingual prompting have the top score for eight out of twelve experiments, with basic multilingual topping the other four.
Multilingual prompting strategies increase reason entropy for social norms questions compared to the best performing diversity increasing comparison methods by a factor of 1.8x-2.38x across all four models, and agreement entropy between 1.65- 2.86x. The demographic diversity increase is more modest, but still consistent, between 1.1-1.2x. 
Impressively, when comparing to the monolingual baseline, multilingual prompting methods can get to up to a 6x increase in reason entropy (LLaMA-70B), 7.3x increase in agreement entropy (LLaMA-8B), and 1.35x (LLaMA-70B) increase in demographic entropies. 

Beyond outperforming comparison methods and baselines, multilingual prompting methods consistently outdo multicultural prompting methods, suggesting the added importance of language in reaching different regions of an LLM's knowledge base.
Interestingly, the added benefit of language varies depending on the level of added cultural cues in the prompts: language is especially helpful when there is less cultural information in the prompt.
Basic multilingual prompting performs markedly better than basic multicultural, by a factor 2x on average for reasoning and agreement entropy (social norm) experiments and 1.1x on average for demographic entropies. Meanwhile, with the exception of two outliers from LLaMA-8B, enhanced multilingual only outperforms enhanced multicultural by a factor of 1.09 on average for reasoning and agreement entropy (social norm) experiments and 1.04x on average demographic entropies. 

Another interesting phenomenon we observe is that occasionally, basic multilingual prompting yields higher demographic entropy than enhanced multilingual prompting. We suggest that this is the result of a narrowing effect where stronger cultural cues for a \emph{specific} cultural background in enhanced multilingual prompting can lead the model to elicit responses centered around that particular region or culture, sometimes reducing overall diversity. By contrast, the basic multilingual prompts are more likely to be a mix of multiple cultures or regions, thus yielding higher entropy. We see support for this idea when analyzing the names generated across the two kinds of prompts. For example, when comparing enhanced and basic multilingual prompting in Chinese, 
GPT-4o produces 547 Chinese names with the enhanced multilingual prompt versus 427 with the basic multilingual prompt out of 1050 names generated. 
Similarly, GPT-4o-mini produced 597 versus 369 Chinese names, and LLaMA-8B produced 728 versus 293 Chinese names in enhanced multilingual versus basic multilingual, respectively. 
This behavior may suggest that basic multilingual prompting can be preferable for achieving broad demographic coverage, whereas enhanced multilingual prompting may be more suitable when culturally specific responses are desired.

Taken together, our results suggest that both linguistic variation and cultural cues in prompts serve as valuable signals for models to generate more inclusive and varied content, reflecting a broader range of perspectives and cultural attributes. 
Further, our results show that using these cultural and language cues together is a more effective strategy for eliciting diverse responses from the model than other diversity-enhancing methods. These results may be surprising given prior work showing minimal impact of language in eliciting \emph{specific} cultural perspectives~\cite{kwok2024evaluating}, but align with prior work suggesting that LLMs have language-specific knowledge bases~\cite{aggarwal2025language}.
In the next section, we show that beyond mild gains in improving diversity, multi\emph{lingual} prompting performs better than multi\emph{cultural} prompting, as we see that 
multi\emph{lingual} prompting prevents hallucination about culturally-relevant information.

 \begin{figure*}[!t]
    \centering
    \includegraphics[width=1\textwidth]{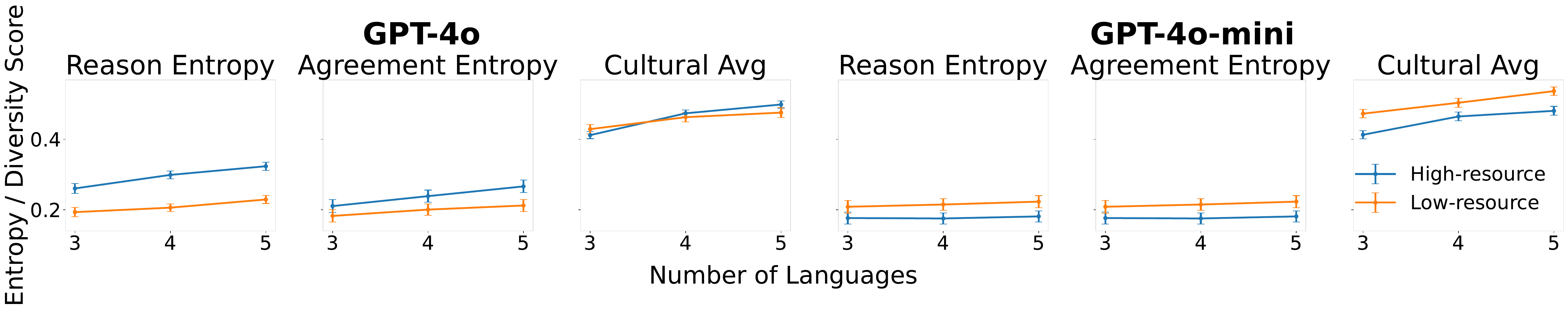}
    \caption{Diversity comparison for GPT-4o and GPT-4o-mini across multilingual methods.}
    \label{fig:diversity-4o}
\end{figure*}




%% file: emilys_version/hallucination.tex
\label{hallucination}
\begin{figure}[t]
    \centering
    \includegraphics[width=0.9\linewidth]{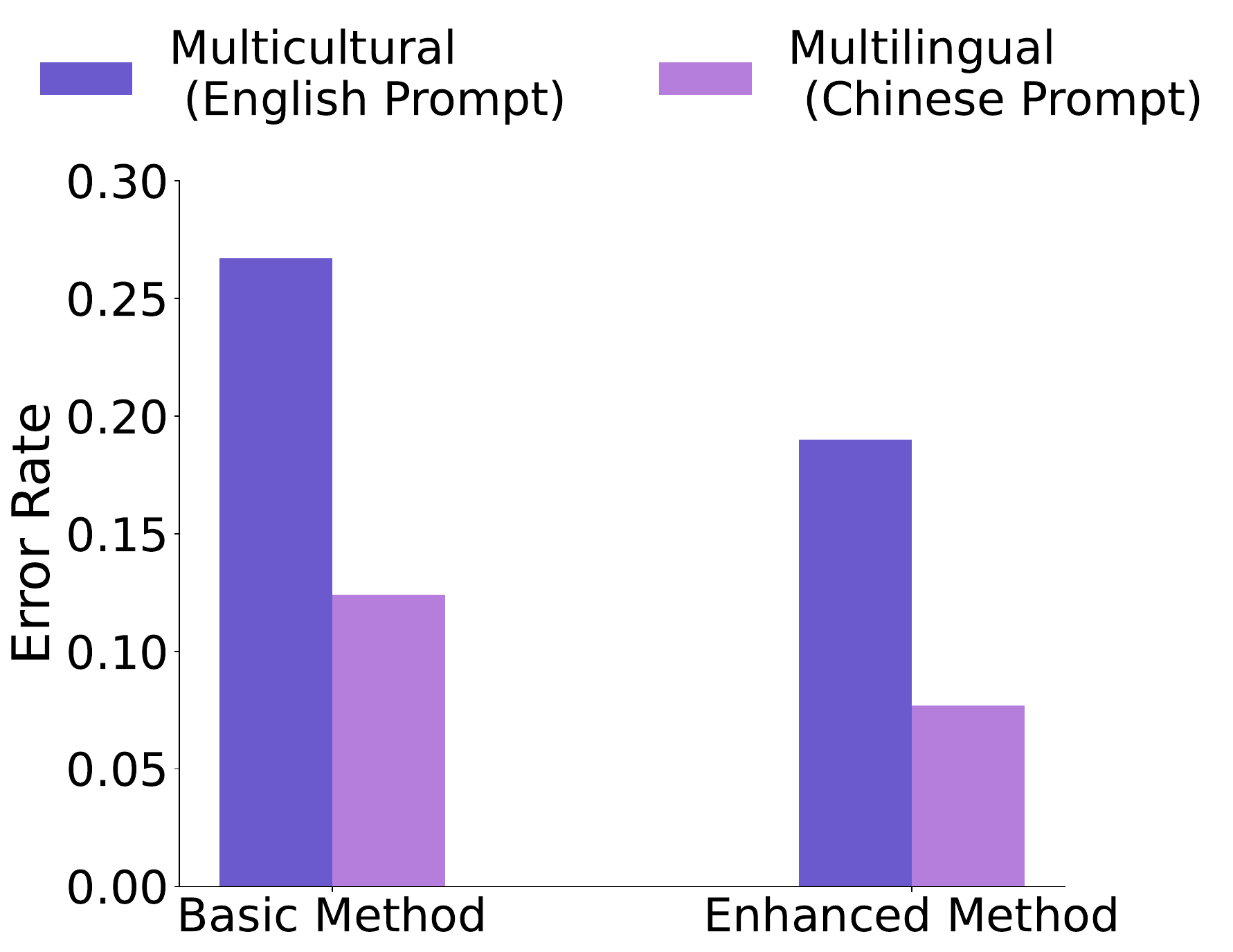}
    \caption{Error rates of Chinese names generated under two prompting strategies. Using multilingual prompts in Chinese yields a lower error rate compared to multicultural prompts (cultural cues but without including the relevant language) in English, demonstrating that prompting in the relevant language reduces hallucination.}
    \label{fig:hallucination}
\end{figure}

We now demonstrate that language is an important component of multilingual prompting, as it can lead to lower hallucination rates for recalling information related to cultures and nationalities where English is not a main language. In particular,
we demonstrate that multicultural prompting with cultural cues but without including the relevant language, (i.e., ``Chinese-speaking, born in Bejing'' but the prompt is not in Chinese) can lead to higher hallucination rates on non-Western names in queries about individuals. For example, in the multicultural setting, when asked, “Who are some circus performers that you admire?”, the model responded with “Zhang Yimou.” However, Zhang Yimou is not a circus performer but rather a renowned Chinese film director. Such errors highlight how excluding the relevant language in prompts can increase hallucinations.


\subsection{Experimental Setup} 
For this experiment, we test hallucination rates on Chinese names generated in response to questions about individuals in various professions.
To do so, we first
randomly sample profession-name pairs generated by the Chinese prompt component of the (basic and enhanced) multilingual and multicultural prompting strategies on the People Diversity Dataset, which asks about naming individuals from different professions. Within the profession-name pairs generated by the Chinese modified prompt, we specifically sample profession-name pairs that were annotated as Chinese by the labeling LLM.
We sample 105 pairs each for the basic multilingual, multicultural, enhanced multilingual, and enhanced multicultural methods. 


Then, to calculate the hallucination rate of generated names, we collected human annotations through Prolific. 
We classify a name as hallucinated for a given profession query if that name is not associated with a person in that profession through a Google or Wikipedia search.
Annotators were given a name and profession from the LLM generation. They are instructed to search the provided 
name on Google and Wikipedia, and report whether the name is likely 
a hallucination i.e., not associated with someone of that profession, or not. 
To ensure annotation
accuracy from Prolific annotators, each name is evaluated independently by three different annotators. Authors manually inspect inconsistent cases (details in Appendix~\ref{appendix:a3}).

\subsection{Results}

\noindent\textbf{Language Helps Prevent Hallucination.} The evaluation reveals a notable difference between the hallucination rate of Chinese names generated from a prompt in Chinese, versus in English.
The multilingual strategy (using prompts in Chinese with Chinese cultural cues) achieves an error rate of 13 out of 105 (12.4\%), whereas the multicultural strategy (using prompts in English with Chinese cultural cues) attains a higher rate of 28 out of 105 (26.7\%). 
This 14\% difference suggests that using the relevant language to signal the model to provide responses about a given culture is an important component of generating diverse responses that are also factually correct. 
Moreover, the enhanced multilingual strategy (using prompts in Chinese with more elaborate Chinese cultural cues) achieves an error rate of 8 out of 105 (7.7\%), whereas the enhanced multicultural strategy (using prompts in English with more elaborate Chinese cultural cues) attains a higher rate of 20 out of 105 (19.0\%). 
These results confirm a trend seen in prior work, which has shown that LLMs 
have different factual knowledge across different languages~\cite{aggarwal2025language}. 



%% file: emilys_version/high_low.tex
\label{sec:high_low}
To further investigate the dynamics of multilingual prompting, we 
test whether diversity gains increase as the number of languages increases, and the performance of the technique across high versus low resource languages. Overall, we find that as the number of languages increases, diversity increases. Interestingly, we find that the performance of multi-lingual prompting across low and high resource languages is model-specific.

\subsection{Experimental Setup}
\label{sec:high_low_setup}
We evaluate two multilingual settings, both of which have English as a base language. One setting adds high-resourced languages for diversity increase: English, Chinese, Japanese, Spanish, and French; and the other setting adds of lower-resourced languages---Nepali, Thai, Turkish, and Ukrainian~\cite{aggarwal2025language}. 
Additionally, we examine how the number of languages used for multilingual prompting (i.e., 3, 4, or 5 languages) affects output diversity, providing insights into whether prompt-level language variety exhibits linear or saturating gains. 

To ensure that our high‑ versus lower‑resourced experiments remain
\emph{directly comparable} across the \(k=3,4,5\) language settings, we standardize both the amount of data collected and the scale on which each diversity metric is reported. Details on how this is done are in Appendix~\ref{appendix:normalization}. 
To ensure that models performed sufficiently well on lower-resourced languages to include in this experiment, we extend our performance check from Section~\ref{sec:experiment_setup} to lower-resourced languages, as well as testing instruction following. Results are in Appendix~\ref{appendix:sanity_check}. GPT-4o and GPT-4o-mini perform well, and LLaMA-70B and 8B do not, so we do not include them.

\subsection{Interaction Effects between Model Size and Resource Level}
\label{sec:high_low_result}
Our results are presented in Figure~\ref{fig:diversity-4o}. 
Overall, we observe that increasing the number of languages from 3 to 5 improves diversity.

Further, our results reveal that diversity performance across low and high resource languages is model-specific. 
For the larger GPT-4o model, high-resourced language combinations consistently yield higher diversity scores across all three metrics—Reason Entropy, Agreement Entropy, and Perspective Diversity. In contrast, for the smaller GPT-4o-mini model, lower-resourced language combinations outperform high-resource ones.

%% file: conclusion.tex
We introduce multilingual and multicultural prompting methods to 
enhance cultural diversity in LLM-generated responses. We show that they outperform existing methods for this task. 
Moreover, we find that multilingual prompting---where the language matches the cultural cues added to each modified version of the original LLM query---is more effective than multicultural prompting---which simply provides cultural cues for various cultures but maintains all modified prompts in English--- both for promoting diversity, and for reducing model hallucination about culture-specific information. This suggests that language \emph{is} an important component in eliciting more diverse responses.
We hope that our method can be an easy, accessible way to increase LLM generation diversity for relevant tasks.

\section{Limitations}

Finally, we discuss some limitations of our work.
Broadly, enhancing diversity may not always be a good outcome. Establishing when is the right time to elicit diverse responses is out of scope for this work, but we look forward to exploring in future work.

Another limitation of our work is that we only explore concatenation as an aggregation strategy--- for tasks which require succinct answers, other aggregation strategies, such as summarizing all answers from multilingual prompt components, or random selection from a distribution of generated responses to different prompt components would be a better choice. While we believe random selection would give identical results in aggregate, fully exploring how to synthesize the diverse perspectives and pieces of information generated through multilingual prompting requires more study, which we look forward to in future work. 

Further, language translation represents another potential limitation. While the authors possess fluency in English, Chinese, and Japanese, translations involving other languages were conducted using GPT models (GPT-4o). 
Existing evaluations and our empirical observations commonly suggest that GPT achieves near-human performance in translation tasks; however, subtle semantic or cultural nuances may not be fully captured in some instances.

Additionally, to ensure reproducibility and reinforce the transparency of our findings, we have included the complete set of prompts and additional experimental outputs in the appendix.
The supplementary materials are intended to facilitate the verification of our results and support the trustworthiness of our conclusions.

\section{Acknowledgments}
We would like to thank Chuhan Ku for the help on data annotation, Falaah Arif Khan for valuable discussions about the evaluation, and Eunsol Choi for the conceptual design.
We would also thank the constructive comments from reviewers.
This material is based upon work supported by the National Science Foundation (NSF) under Grant No. IIS-2239862.

%% file: appendix.tex
\section*{Appendix Roadmap}
This appendix provides supplementary details supporting our main paper. 
Appendix~\ref{Prompts_of_Social_Norms_Experiment} and ~\ref{Prompts_for_Cultural_Representation_Experiments} describes the exact prompts used in the social norms 
and cultural representation experiments. Appendix~\ref{apppendix:annotation_demographic} outlines 
our annotation procedure for demographic attributes of generated names. 
Appendix~\ref{appendix:normalization} and  ~\ref{name_normalization} explains the normalization and calculation of our 
diversity metrics. Appendix~\ref{Detailed_Results} presents full experimental results 
(including tables and figures) that extend the main text. 
Appendix~\ref{appendix:sanity_check} and ~\ref{wvs} details 
our performance and validity checks. Appendix ~\ref{appendix:full_result} provide additioanl result and ~\ref{comp_cost} discuss the computational cost of different methods.

\subsection{Prompts of Social Norms Experiment}
\label{Prompts_of_Social_Norms_Experiment}
This section contains the prompts used in social norms experiment.

\subsubsection{Generating Choices}
\label{generating_choices}
All multi-choice questions are constructed using statements from the Social-Chem-101 dataset~\cite{forbes2020social}, a publicly available opensource dataset designed to reflect diverse human moral judgments. The dataset is licensed under the CC BY-SA 4.0 license, which can be adapted for any purposes as long as we follow the Attribution and ShareAlike terms (and we do follow them).

To generate a range of perspectives on social norm statements, we used the following prompt template with the GPT-4o model:

\begin{quote}
\small
\texttt{Given the following statement: "\{statement\}"\\
Provide four different perspectives (two agreeing and two disagreeing).\\
Each perspective should include:\\
- A valence: "agree" or "disagree"\\
- An explanation for why someone might hold this view.\\
\\
Example response:\\
Statement: You should not feel obligated to go to your family reunions\\
1. \{ "valence": "agree", "explanation": "Obligations shouldn't dictate personal choices; you should only go if you want to." \}\\
2. \{ "valence": "agree", "explanation": "Forced interactions can cause unnecessary stress and discomfort." \}\\
3. \{ "valence": "disagree", "explanation": "Family reunions are important for maintaining family bonds." \}\\
4. \{ "valence": "disagree", "explanation": "Spending time with family, even if you don’t remember them well, can create new memories." \}\\
Provide your response in JSON format.}\\
\end{quote}

\vspace{0.5em}
We show two representative outputs in Table~\ref{tab:rot-examples}, each including two agreeing and two disagreeing perspectives.
A potential avenue for future research in opinion generation involves the incorporation of Retrieval-Augmented Generation (RAG) techniques, which may enhance the grounding of generated opinions in identifiable sociocultural frameworks (e.g., Confucian ethics, modern family structures).

\begin{table*}[t]
\centering
\small
\renewcommand{\arraystretch}{1.4}
\begin{tabular}{p{0.95\textwidth}}
\hline
\textbf{Statement:} \textit{“It is wrong to do drugs.”} \\
\hline
A: Agree. Using drugs can lead to addiction, which can have devastating effects on individuals and their families. \\
B: Agree. Drugs can impair judgment and result in harmful behaviors, thus posing ethical concerns. \\
C: Disagree. Individuals should have the freedom to make personal choices about drug use, provided they don’t harm others. \\
D: Disagree. In some contexts, drugs serve spiritual or medicinal purposes, and thus their use may not be universally wrong. \\
\hline
\end{tabular}

\vspace{1em}

\begin{tabular}{p{0.95\textwidth}}
\hline
\textbf{Statement:} \textit{“It's not okay to spend money on things you could do yourself.”} \\
\hline
A: Agree. Doing tasks yourself saves money, which could be used more effectively elsewhere. \\
B: Agree. Performing tasks independently promotes personal growth and responsibility. \\
C: Disagree. Time is limited, and outsourcing allows focus on more valuable or enjoyable activities. \\
D: Disagree. Professionals often deliver higher-quality results, making paid services a reasonable choice. \\
\hline
\end{tabular}
\caption{Examples of model-generated perspectives on social norm statements. Each includes two agreeing and two disagreeing viewpoints labeled A–D.}
\label{tab:rot-examples}
\end{table*}

\subsubsection{Social Norm Questions}
\label{prompt_socialnorm}
\begin{figure*}[!t]
    \centering
    \includegraphics[width=1\textwidth]{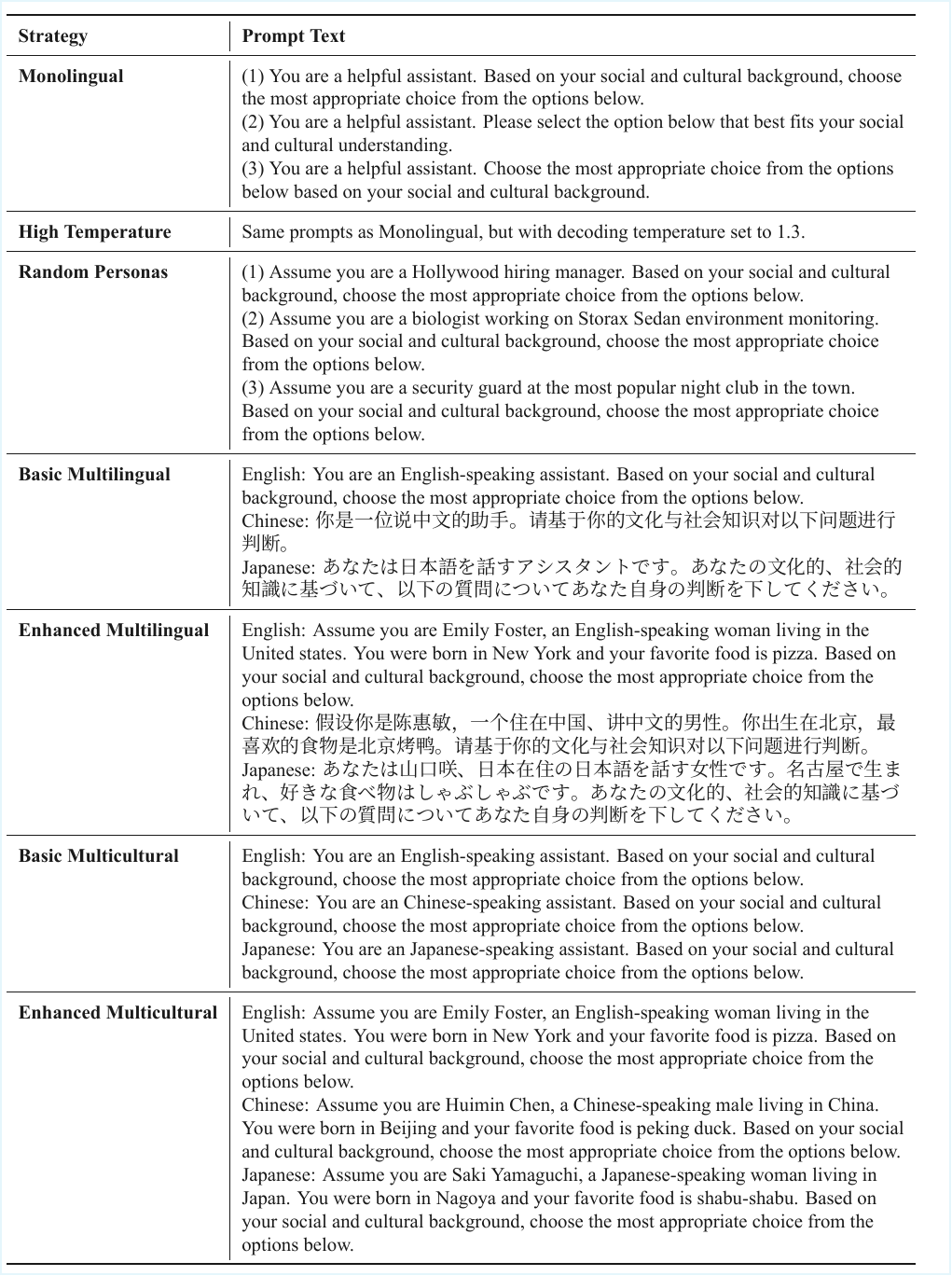}
    \vspace{20pt}
    \caption{Prompts for social norm questions}
    \label{fig:socialchem_prompt}
\end{figure*}

Prompts we use for social norm questions are shown Figure~\ref{fig:socialchem_prompt}.
To ensure a fair comparison across strategies, we standardize the number and variety of prompts used. 
For the Monolingual strategy and High Temperature strategy, we paraphrase the same base prompt into three distinct versions. 
In the Random Personas strategy, we initially generate 10 diverse persona-based prompts and randomly select 3 for use. 
For both Enhanced-Multilingual and Enhanced-Multiculture conditions, we first create 10 unique prompts for each language and then randomly select one prompt per language to include in the evaluation. 
This approach maintains consistency in prompt diversity while allowing us to test the effects of linguistic and persona variation.

\subsection{Prompts for Cultural Representation Experiments}
\label{Prompts_for_Cultural_Representation_Experiments}

\subsubsection{Cultural Representation Questions}
This section contains the prompts used in the cultural representation experiment. All questions for this experiment are from an open source dataset called the People Diversity Dataset~\cite{lahoti2023improving}, which includes 105 profession-related prompts such as 'Name some writers who inspire you' or 'Who are your favorite singerscientists?'. We used these questions directly and asked the LLMs to generate exactly 10 names for each prompt.  
This dataset and its paper are licensed under the CC BY-SA 4.0 license, which can be adapted for any purposes as long as we follow the Attribution and ShareAlike terms (and we do follow them).

\begin{figure*}[!t]
    \centering
    \includegraphics[width=1\textwidth]{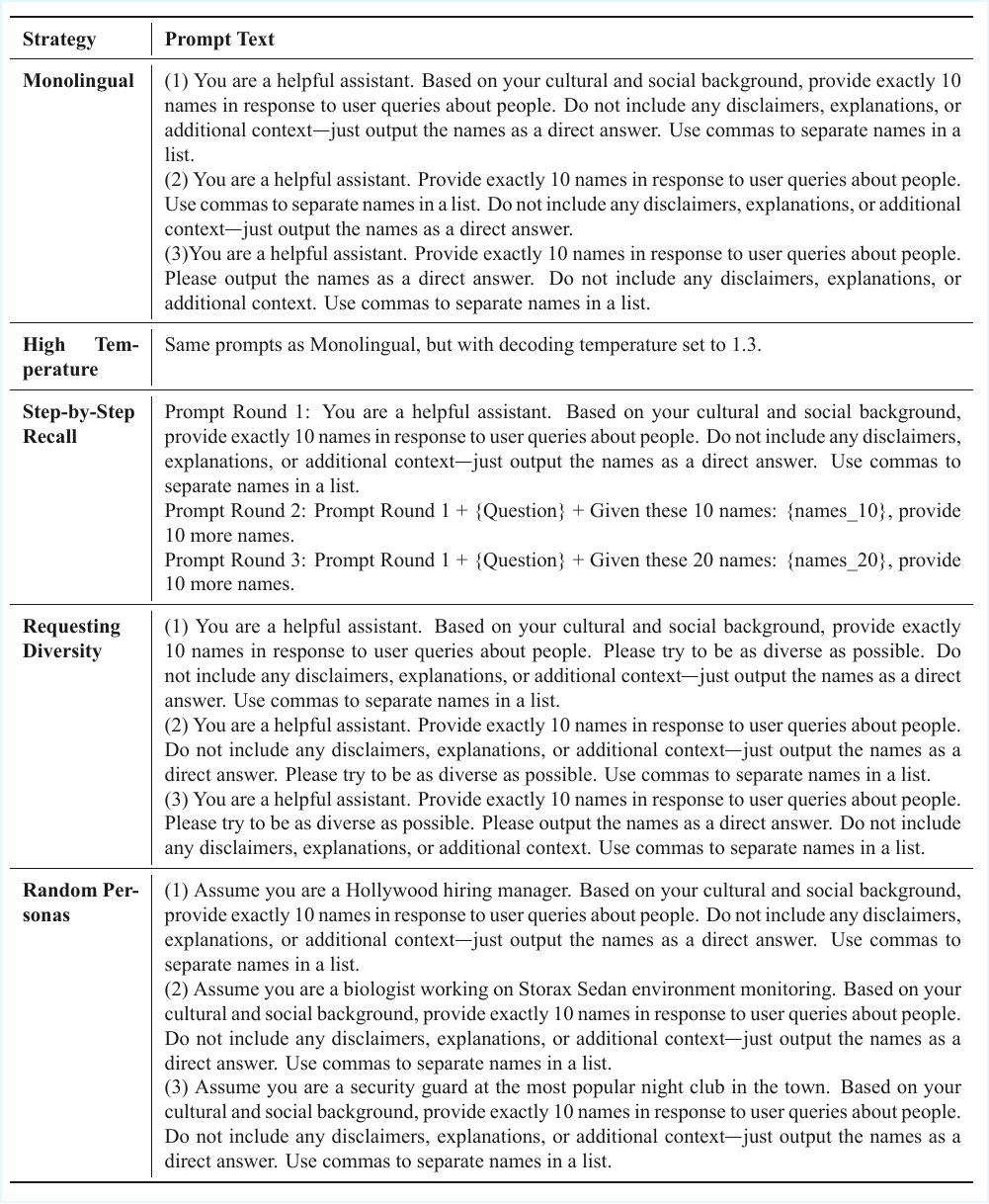}
    \vspace{-10pt}
    \caption{Prompts for cultural representation questions - baseline and other diversity-enhancing methods}
    \label{fig:culture_prompt1}
\end{figure*}

\begin{figure*}[!t]
    \centering
    \includegraphics[width=1\textwidth]{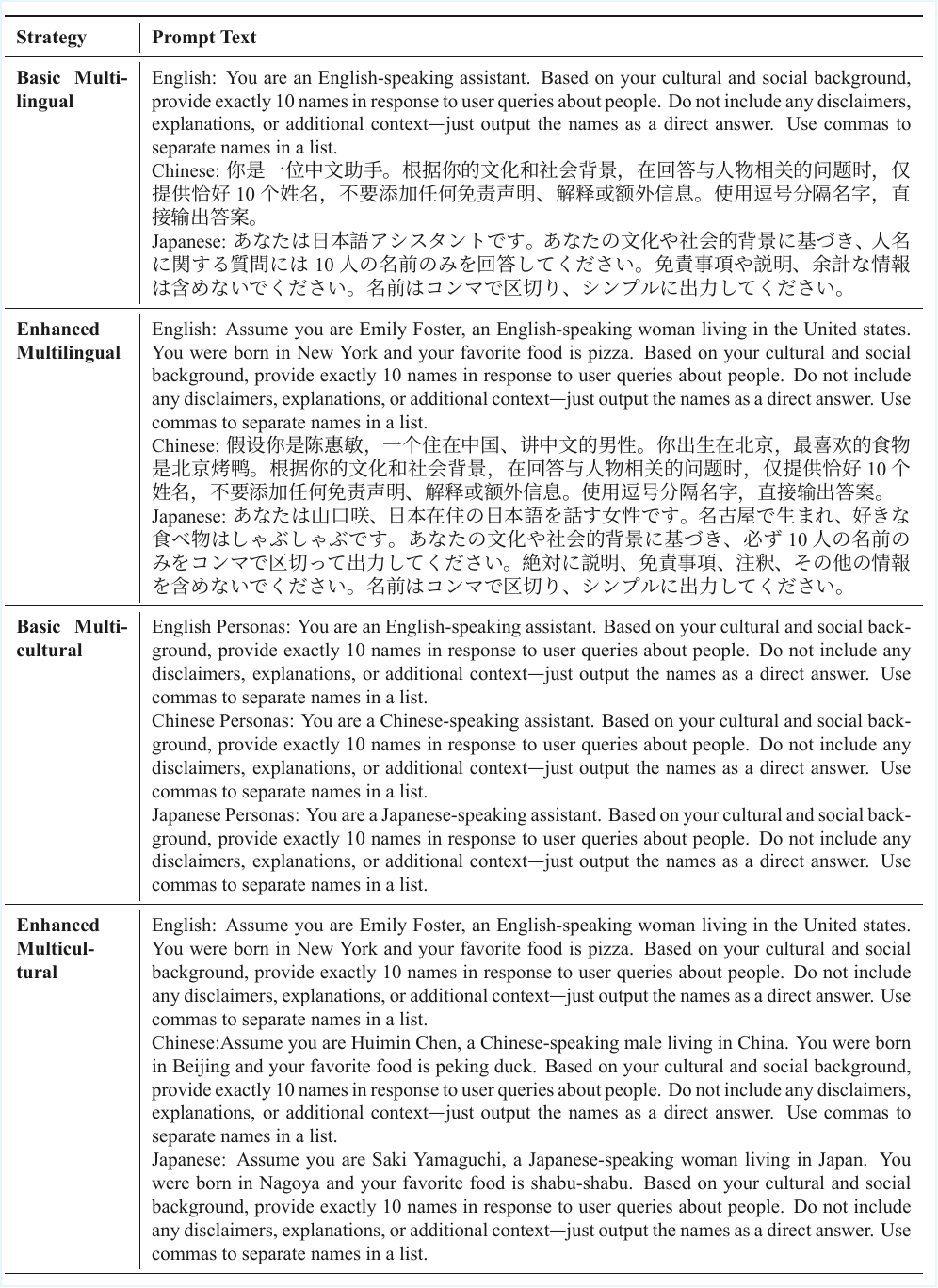}
    \vspace{15pt}
    \caption{Prompts for cultural representation questions - our multilingual and multiculture strategies}
    \label{fig:culture_prompt2}
\end{figure*}

\label{prompt_culture}
Prompts we use for cultural representation questions are shown
in Figure~\ref{fig:culture_prompt1} and Figure~\ref{fig:culture_prompt2}. For each question, we run all three prompts under each condition (e.g., the three paraphrased English prompts for the Monolingual condition, or the three language-specific prompts for the Multilingual condition). Each prompt elicits 10 name responses, resulting in a total of 30 names per condition. We then annotate each name with its likely nationality, ethnicity, and region. To quantify the diversity of cultural perspectives generated by each prompting strategy, we compute the entropy across the 30 annotated names.

\subsubsection{Annotation for Demographic Details of Names}
\label{apppendix:annotation_demographic}
To estimate the cultural and geographic origin of names, we used the following prompt:
\begin{quote}
\small
\texttt{
Given the name "\{name\}", predict the following: \\
- Likely nationality or country of origin \\
- Likely ethnic or cultural background \\
- Likely region (Choose ONLY from: Europe, East Asia, South Asia, Middle East, North America, Latin America, Africa, Oceania) \\
\\
Provide the result in a structured JSON format: \\
\{ \\
\ \ \ "nationality": "...", \\
\ \ \ "ethnicity": "...", \\
\ \ \ "region": "..." \\
\}
}
\end{quote}

\begin{table}[t]
\centering
\small
\renewcommand{\arraystretch}{1.4}
\begin{tabular}{p{0.95\linewidth}}
\hline
\textbf{Name:} \texttt{Galileo} \\
\texttt{\{ "nationality": "Italian", "ethnicity": "Italian", "region": "Europe" \}} \\
\hline
\textbf{Name:} \texttt{Yao Ming} \\
\texttt{\{ "nationality": "Chinese", "ethnicity": "Han Chinese", "region": "East Asia" \}} \\
\hline
\end{tabular}
\caption{Examples of cultural annotations predicted for given names.}
\label{tab:culture-annot-examples}
\end{table}

\vspace{0.5em}
Table~\ref{tab:culture-annot-examples} shows two illustrative examples.
To ensure the reliability of these annotations, we conduct 20 manual sanity checks for each prompting strategy. For each check, we verify whether the predicted nationality, ethnicity, and region are reasonable given the input name. 
Results show that the majority of outputs aligned well with publicly known information about the names. Overall, the annotation accuracy across strategies is approximately 90\%.

\subsection{Metric Normalization for Social Norm Experiemnt}
\label{appendix:normalization}
Let \(k\) be the number of model answers collected for the same statement  
(\(k \in \{3,4,5\}\) in our experiments) and let \(m\) be the number of
mutually–exclusive categories used by the metric  
(\(m=4\) for \textbf{Reason}, \(m=2\) for \textbf{Valence}).
We rescale every raw score \(H\) to the interval \([0,1]\) via its
\emph{theoretical upper bound} \(H_{\max}(k,m)\):
\[
\widetilde{H}(k,m)=\frac{H}{H_{\max}(k,m)}.
\]

\paragraph{General form of \(H_{\max}(k,m)\).}
Entropy is maximized when the \(k\) answers are spread as evenly as possible
across the \(m\) categories.  Write
\[
q = \Bigl\lfloor \frac{k}{m} \Bigr\rfloor,\qquad
r = k - mq \quad (0 \le r < m),
\]
so that \(r\) categories receive \(q+1\) answers and the remaining \(m-r\)
categories receive \(q\) answers.  The corresponding empirical probabilities
are
\[
p_{\text{high}} = \frac{q+1}{k}, \qquad
p_{\text{low}}  = \frac{q}{k},
\]
\noindent\textbf{Maximal entropy.}
Let \(p_h=(q+1)/k\) and \(p_\ell=q/k\).
Then
\[
  H_{\max}(k,m)=
  -\,r\,p_h\log p_h
  -\,(m-r)\,p_\ell\log p_\ell .
\]
(We adopt the convention \(0\log 0 := 0\) whenever a probability is zero.)
\medskip
\begin{itemize}[leftmargin=*]
  \item \textbf{Reason Entropy} (\(m=4\)):
  \[
  \begin{aligned}
  k=3 &:~ H_{\max} = \log 3,\\
  k=4 &:~ H_{\max} = \log 4,\\
  k=5 &: H_{\max} =
         -\!\Bigl(\tfrac25\log\tfrac25
                  + \,\tfrac35\log\tfrac15\Bigr)
         \approx 1.332.
  \end{aligned}
  \]

  \item \textbf{Valence Entropy} (\(m=2\)):
  \[
  \begin{aligned}
  k=3 &:~ H_{\max}=-\!\bigl(\tfrac13\log\tfrac13+\tfrac23\log\tfrac23\bigr)\approx0.637,\\
  k=4 &:~ H_{\max}=\log 2\approx0.693,\\
  k=5 &:~ H_{\max}=-\!\bigl(\tfrac25\log\tfrac25+\tfrac35\log\tfrac35\bigr)\approx0.673.
  \end{aligned}
  \]

  \item \textbf{Perspective Diversity (a.k.a.\ Perspective Entropy).}  
        For each statement we embed the four choices
        \(\mathcal{E}=\{\mathbf e_A,\mathbf e_B,\mathbf e_C,\mathbf e_D\}\)
        using Sentence‑BERT.  
        With \(k\) languages (\(k\!\in\!\{3,4,5\}\)), consider every
        size‑\(k\) subset \(S\subseteq\mathcal{E}\).
        For any subset \(S=\{i_1,\dots,i_k\}\) we define its mean pairwise
dissimilarity
\[
  D(S)\;=\;
  \frac{2}{k(k-1)}
  \sum_{a<b}
  \Bigl[
      1-
      \frac{\mathbf e_{i_a}\!\cdot\!\mathbf e_{i_b}}
           {\lVert\mathbf e_{i_a}\rVert\,\lVert\mathbf e_{i_b}\rVert}
  \Bigr].
\]

        For the same statement \(q\) we set its empirical upper bound to
        \begin{equation}
            H_{\max}^{(q)}(k)
            = \max_{\substack{S\subseteq\mathcal E\\|S|=k}} D^{(q)}(S),
        \end{equation}
        i.e.\ the largest dissimilarity obtainable from any size‑\(k\) subset.
        
        \textit{Example (\(k=3\)).}  
        The four triplets
        \(ABC, ABD, ACD, BCD\) are evaluated; assume the maximum is
        \(D(ACD)\).  
        If the model produced the labels \(ABD\), then for this statement
        \(\widetilde H_{\text{Persp}}(q,3)=D(ABD)/D(ACD)\).

        Averaging \(\widetilde H_{\text{Persp}}(q,k)\) over all statements
        places the metric on the common \([0,1]\) scale:
        \(1\) indicates the greatest possible diversity, \(0\) indicates
        none.

After normalization, every metric lies on the same \([0,1]\) scale:  
\(\widetilde{H}=1\) denotes the greatest possible diversity,  
while \(\widetilde{H}=0\) indicates none.

\end{itemize}

\subsection{Metric Normalization for Cultural Representation Experiment}
\label{name_normalization}
To place the cultural–diversity metrics on a common \([0,1]\) scale we
again rescale each raw entropy score \(H\) by its theoretical upper bound
\(H_{\max}\):
\[
\widetilde H \;=\; \frac{H}{H_{\max}}.
\]

\begin{itemize}
  \item \textbf{Nationality \& Ethnicity.}  
        For every question we collect exactly \(k=30\) names, regardless of
        the number of languages.  The largest entropy occurs when all
        \(30\) names belong to distinct categories, giving
        \[
            H_{\max}= \log 30.
        \]

  \item \textbf{Region.}  
        The attribute “region” has \(m=8\) possible categories.  
        Spreading the same \(k=30\) names as evenly as possible across those
        eight categories maximises the entropy.  
        With
        \(
            q=\bigl\lfloor \tfrac{k}{m} \bigr\rfloor = 3
        \)
        and
        \(r = k - mq = 6\),
        six regions receive \(q+1 = 4\) names and the remaining two receive
        \(q=3\).
        Setting \(p_h=\tfrac{4}{30}\) and \(p_\ell=\tfrac{3}{30}\) we obtain
        \[
            H_{\max} = -\,6\,p_h\log p_h - 2\,p_\ell\log p_\ell .
        \]
\end{itemize}

After this normalization every metric lies in \([0,1]\);  
\(\widetilde H=1\) denotes the greatest possible diversity under the
30‑name constraint, while \(\widetilde H=0\) indicates none.

\subsection{Detailed Results of Demographic and Social Norm Experiments}
\label{Detailed_Results}
This section provides the complete results of Table~\ref{tab:merged_diversity_final} in the main paper.
Table~\ref{tab:social_updated_grouped} presents the full results of the social norm experiment, reporting diversity metrics across prompting strategies and models. Table~\ref{tab:name-updated} presents the full results of the cultural representation experiment.

\definecolor{lightpurple}{RGB}{230, 230, 250}

\begin{table*}
\centering
\resizebox{0.8\textwidth}{!}{%
\begin{tabular}{llccc}
\toprule
\textbf{Model} & \textbf{Strategy} & \textbf{Reason} & \textbf{Agreement} & \textbf{Perspective} \\

\midrule
\multirow{9}{*}{GPT-4o}
 & \cellcolor{gray!15}\textbf{Baseline} & \cellcolor{gray!15} & \cellcolor{gray!15} & \cellcolor{gray!15} \\
 & Monolingual & 0.079 & 0.076 & 0.077 \\
 & \cellcolor{gray!15}\textbf{Diversity-Enhancing} & \cellcolor{gray!15} & \cellcolor{gray!15} & \cellcolor{gray!15} \\
 & High Temperature & 0.161 & 0.128 & 0.158 \\
 & Random Personas & 0.166 & 0.150 & 0.167 \\
 & \cellcolor{gray!15}\textbf{Our Methods} & \cellcolor{gray!15} & \cellcolor{gray!15} & \cellcolor{gray!15} \\
 & Basic Multicultural & 0.191 & 0.172 & 0.192 \\
& Basic Multilingual & 0.249 & 0.210 & 0.240 \\
 & Enhanced Multicultural & 0.280 & 0.245 & 0.273 \\
 & Enhanced Multilingual & \textbf{0.300} & \textbf{0.247} & \textbf{0.295} \\

\midrule
\multirow{9}{*}{GPT-4o-mini}
 & \cellcolor{gray!15}\textbf{Baseline} & \cellcolor{gray!15} & \cellcolor{gray!15} & \cellcolor{gray!15} \\
 & Monolingual & 0.089 & 0.050 & 0.085 \\
 & \cellcolor{gray!15}\textbf{Diversity-Enhancing} & \cellcolor{gray!15} & \cellcolor{gray!15} & \cellcolor{gray!15} \\
 & High Temperature & 0.121 & 0.058 & 0.114 \\
 & Random Personas & 0.128 & 0.088 & 0.129 \\
 & \cellcolor{gray!15}\textbf{Our Methods} & \cellcolor{gray!15} & \cellcolor{gray!15} & \cellcolor{gray!15} \\
 & Basic Multicultural  & 0.127 & 0.096 & 0.123 \\
  & Basic Multilingual & 0.299 & 0.176 & 0.292 \\
 & Intense Multicultural & 0.167 & 0.102 & 0.162 \\
 & Intense Multilingual & \textbf{0.304} & \textbf{0.190} & \textbf{0.298} \\

\hline
\multirow{9}{*}{LLaMA 70B}
 & \cellcolor{gray!15}\textbf{Baseline} & \cellcolor{gray!15} & \cellcolor{gray!15} & \cellcolor{gray!15} \\
 & Monolingual & 0.050 & 0.048 & 0.051 \\
 & \cellcolor{gray!15}\textbf{Diversity-Enhancing} & \cellcolor{gray!15} & \cellcolor{gray!15} & \cellcolor{gray!15} \\
 & High Temperature & 0.068 & 0.056 & 0.067 \\
 & Random Personas & 0.135 & 0.122 & 0.130 \\
 & \cellcolor{gray!15}\textbf{Our Methods} & \cellcolor{gray!15} & \cellcolor{gray!15} & \cellcolor{gray!15} \\
 & Basic Multicultural & 0.105 & 0.086 & 0.109 \\
  & Basic Multilingual & 0.262 & 0.218 & 0.263 \\
 & Enhanced Multicultural & 0.280 & 0.170 & 0.260 \\
 & Enhanced Multilingual & \textbf{0.304} & \textbf{0.222} & \textbf{0.294} \\

\midrule
\multirow{9}{*}{LLaMA 8B}
 & \cellcolor{gray!15}\textbf{Baseline} & \cellcolor{gray!15} & \cellcolor{gray!15} & \cellcolor{gray!15} \\
 & Monolingual & 0.094 & 0.064 & 0.085 \\
 & \cellcolor{gray!15}\textbf{Diversity-Enhancing} & \cellcolor{gray!15} & \cellcolor{gray!15} & \cellcolor{gray!15} \\
 & High Temperature & 0.236 & 0.164 & 0.225 \\
 & Random Personas & 0.143 & 0.086 & 0.135 \\
 & \cellcolor{gray!15}\textbf{Our Methods} & \cellcolor{gray!15} & \cellcolor{gray!15} & \cellcolor{gray!15} \\
 & Basic Multicultural & 0.257 & 0.208 & 0.247 \\
  & Basic Multilingual & \textbf{0.555} & 0.465 & \textbf{0.529} \\
 & Enhanced Multicultural & 0.164 & 0.070 & 0.150 \\
 & Enhanced Multilingual & 0.471 & \textbf{0.469} & 0.445 \\

\Xhline{1.2pt}
\end{tabular}
}
\caption{Diversity metrics across prompting strategies and models. Bold indicates the highest value within each model. Purple highlight shows the maximum across all models for each metric.}
\label{tab:social_updated_grouped}
\end{table*}

\definecolor{lightpurple}{RGB}{230, 230, 250}

\begin{table*}[t]
\centering
\resizebox{0.7\textwidth}{!}{%
\begin{tabular}{llcccc}
\toprule
\textbf{Model} & \textbf{Strategy} & \textbf{Nationality} & \textbf{Ethnicity} & \textbf{Region} & \textbf{Avg} \\

\midrule
\multirow{12}{*}{GPT-4o}
 & \textbf{Baseline} &  & & & \\
 & Monolingual & 0.335 & 0.421 & 0.190 & 0.315 \\
\rowcolor{gray!15}
 & \textbf{Diversity-Enhancing} & & & & \\
 & Requesting Diversity & 0.378 & 0.482 & 0.250 & 0.370 \\
 & High Temperature & 0.374 & 0.452 & 0.206 & 0.344 \\
 & Step-By-Step Recall & 0.408 & \textbf{0.519} & 0.208 & 0.378 \\
 & Random Personas & 0.351 & 0.450 & 0.202 & 0.335 \\
\rowcolor{gray!15}
 & \textbf{Our Methods} & & & & \\
 & Basic Multicultural & 0.386 & 0.456 & 0.240 & 0.360 \\
 & Basic Multilingual & \textbf{0.465} & 0.500 & \textbf{0.281} & \textbf{0.415} \\
 & Enhanced Multicultural & 0.398 & 0.490 & 0.246 & 0.378 \\
 & Enhanced Multilingual & 0.441 & 0.462 & 0.249 & 0.384 \\
 
\midrule
\multirow{12}{*}{GPT-4o-mini}
 & \textbf{Baseline} & & & & \\
 & Monolingual & 0.322 & 0.429 & 0.189 & 0.314 \\
\rowcolor{gray!15}
 & \textbf{Diversity-Enhancing} & & & & \\
 & Diverse Prompt & 0.356 & 0.465 & 0.227 & 0.349 \\
 & High Temperature & 0.368 & 0.460 & 0.206 & 0.345 \\
 & Step-By-Step Recall & 0.382 & 0.505 & 0.202 & 0.363 \\
 & Random Personas & 0.355 & 0.461 & 0.200 & 0.338 \\
\rowcolor{gray!15}
 & \textbf{Our Methods} & & & & \\
 & Basic Multicultural & 0.421 & 0.466 & \textbf{0.321} & 0.402 \\
  & Basic Multilingual & 0.466 & \textbf{0.516} & 0.295 & 0.426 \\
 & Enhanced Multicultural & 0.442 & 0.474 & 0.254 & 0.390 \\
 & Enhanced Multilingual & \textbf{0.471} & 0.509 & 0.258 & \textbf{0.413} \\

\midrule
\multirow{12}{*}{LLaMA 70B}
 & \textbf{Baseline} & & & & \\
 & Monolingual & 0.335 & 0.411 & 0.188 & 0.311 \\
\rowcolor{gray!15}
 & \textbf{Diversity-Enhancing} & & & & \\
 & Diverse Prompt & 0.353 & 0.458 & 0.212 & 0.341 \\
 & High Temperature & 0.379 & 0.454 & 0.239 & 0.357 \\
 & Step-By-Step Recall & 0.391 & 0.438 & 0.249 & 0.359 \\
 & Random Personas & 0.330 & 0.429 & 0.177 & 0.312 \\
\rowcolor{gray!15}
 & \textbf{Our Methods} & & & & \\
 & Basic Multicultural & 0.416 & 0.429 & 0.287 & 0.377 \\
  & Basic Multilingual & 0.460 & 0.485 & 0.262 & 0.402 \\
 & Enhanced Multicultural & 0.444 & 0.500 & 0.281 & 0.409 \\
 & Enhanced Multilingual & \textbf{0.472} & \textbf{0.520} & \textbf{0.293} & \textbf{0.428} \\

\midrule
\multirow{12}{*}{LLaMA 8B}
 & \textbf{Baseline} & & & & \\
 & Monolingual & 0.351 & 0.435 & 0.189 & 0.325 \\
\rowcolor{gray!15}
 & \textbf{Diversity-Enhancing} & & & & \\
 & Diverse Prompt & 0.345 & 0.433 & 0.188 & 0.322 \\
 & High Temperature & --- & --- & --- & --- \\
 & Step-By-Step Recall & 0.421 & 0.507 & 0.202 & 0.377 \\
 & Random Personas & 0.352 & 0.451 & 0.198 & 0.334 \\
\rowcolor{gray!15}
 & \textbf{Our Methods} & & & & \\
 & Basic Multicultural & 0.429 & 0.464 & 0.249 & 0.380 \\
  & Basic Multilingual & \textbf{0.490} & \textbf{0.509} & \textbf{0.282} & \textbf{0.427} \\
 & Enhanced Multicultural & 0.430 & 0.467 & 0.250 & 0.382 \\
 & Enhanced Multilingual & 0.447 & 0.475 & 0.242 & 0.388 \\
\bottomrule
\end{tabular}
}%
\caption{Normalized cultural diversity scores across prompting strategies and models. \textbf{Avg} is the average of Nationality, Ethnicity, and Region. Bold values indicate the highest score per model.}
\label{tab:name-updated}
\end{table*}

\subsection{Result: Multilingual Prompting Preserves Factual Accuracy}
\label{sec:factuality}
To verify that multilingual prompting does not compromise the factual accuracy of language models, we evaluate their performance on the Multilingual Grade School Math Benchmark {(MGSM)~\cite{shi2022language}}, which consists of mathematical reasoning tasks translated into multiple languages. 

\begin{figure}[t]
    \centering
    \includegraphics[width=\linewidth]{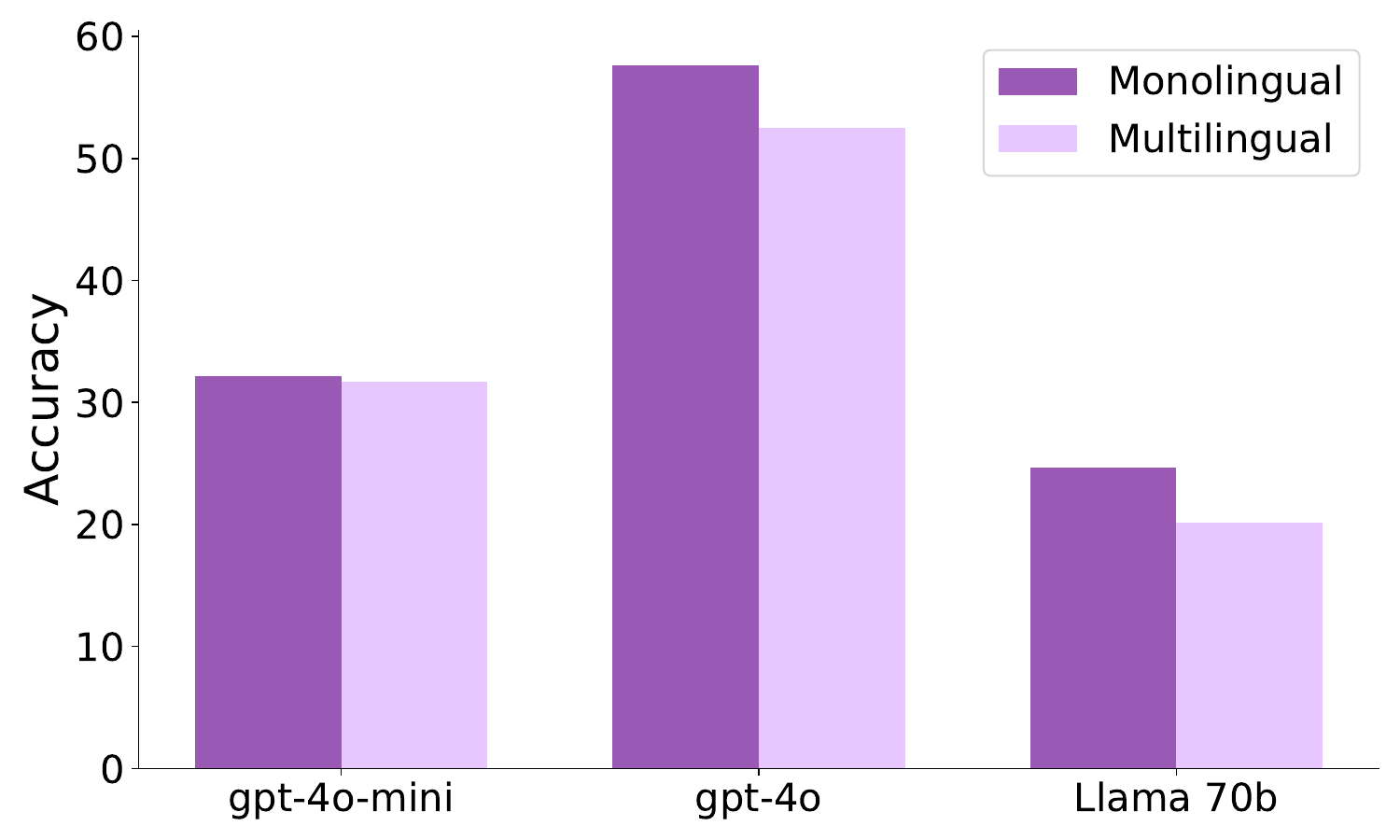}
    \caption{Performance on multilingual grade school math benchmark}
    \label{fig:mgsm}
\end{figure}

Figure~\ref{fig:mgsm} presents the factuality accuracy across three models—GPT-4o-mini, GPT-4o, and LLaMA 70B—under monolingual and multilingual prompting conditions. Across all models, we observe that multilingual prompting maintains comparable factual accuracy to monolingual prompting. GPT-4o-mini shows virtually no change. For GPT-4o and LLaMA-70B, there is a slight  performance drop around 5\%, but the overall competency of the model remains intact.

\subsection{Details of the Human Study}
\label{appendix:a3}
We randomly sample 105 (10\% of the answer) question–name pairs for each from the outputs generated by the Basic Multilingual, Basic Multiculture, Enhanced Multilingual and Enhanced Multiculture strategies under the Chinese language condition. Hence, there are 420 QA Pairs to be annotated in total.

We conduct a human annotation study to evaluate name-based cultural appropriateness using crowd-sourced annotators on Prolific. The study was open to 79,169 eligible participants from a larger Prolific population of 232,330. A total of 420 names were annotated in this study. We recruit 84 annotators from the U.S.-based Prolific participant pool, each of whom annotate 15-16 unique names. Each name is thus evaluated independently by three different annotators to ensure redundancy and allow for inter-rater comparison.

The annotation is conducted through a Google Forms survey, which require no software installation and is accessible via mobile, tablet, or desktop. Custom screening is applied to ensure annotators are fluent in English and located in the United States. Participants are instructed to judge whether the provided name is a reasonable and appropriate answer to a given question. They are asked to verify it using external resources such as Google or Wikipedia and are explicitly instructed not to guess or answer randomly.

Compensation is set at \$2 per participant, equivalent to \$12.00/hour, which is recommended amount by Prolific. The median completion time is approximately 7 minutes. Upon submission, each response is manually reviewed, and a completion code is provided for payment processing. 
The study is classified as exempt by the IRB of authors' institution.

\subsection{Additional Results}
\label{appendix:full_result}
\begin{figure*}[!t]
    \centering
    \includegraphics[width=1\textwidth]{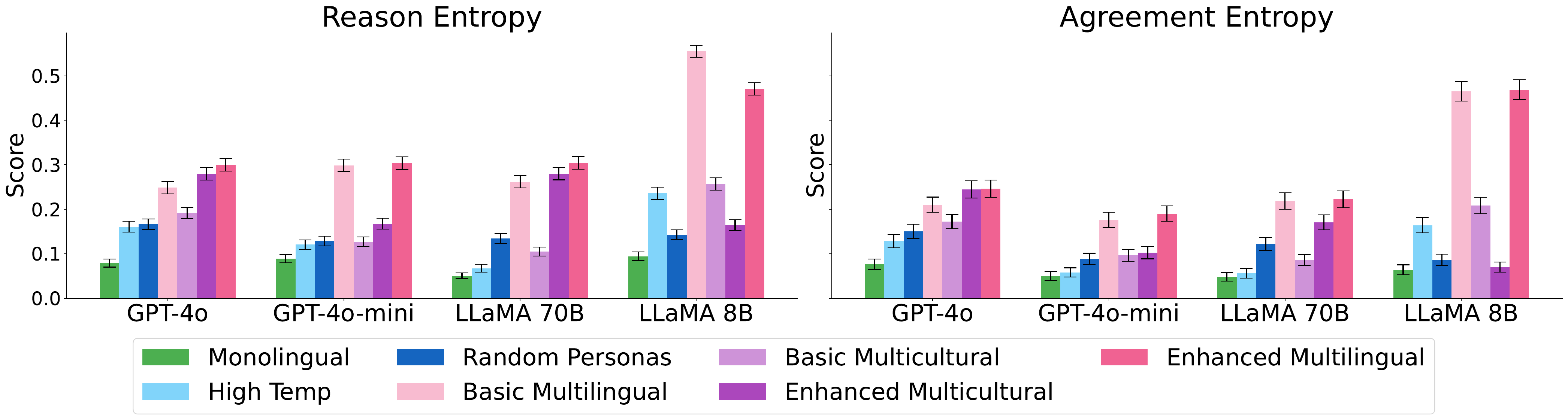}
    \caption{Results of social norm experiment}
    \label{fig:socialchem_plot}
\end{figure*}

\begin{figure*}[!t]
    \centering
    \includegraphics[width=0.9\textwidth]{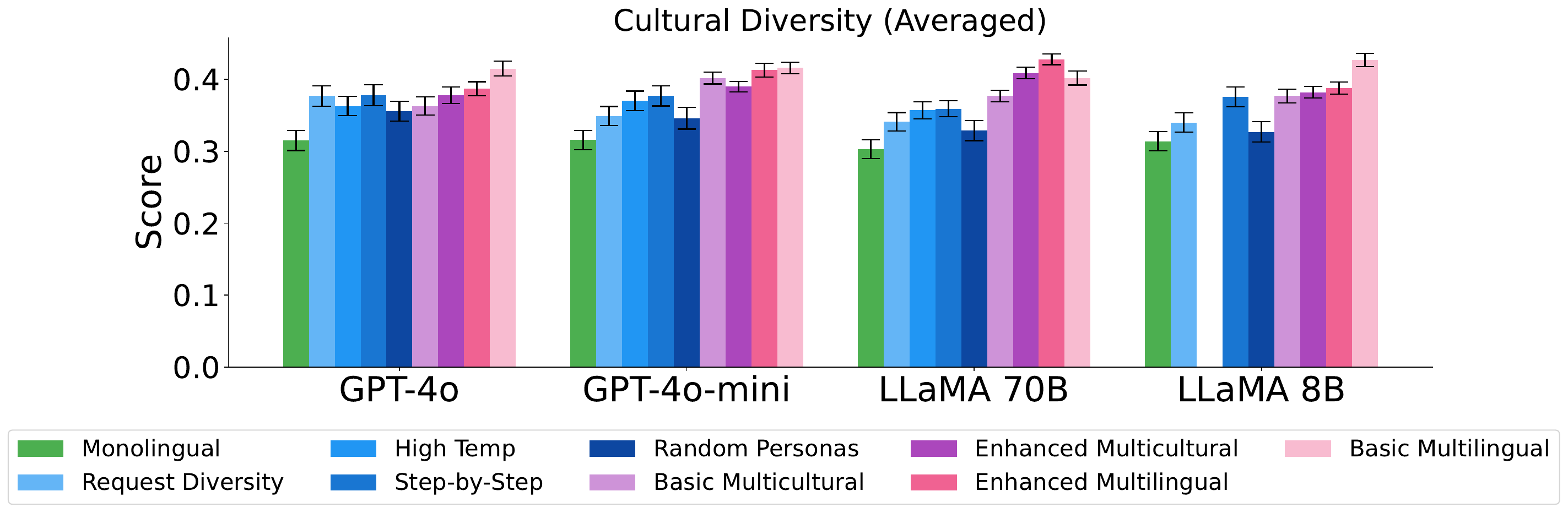}
    \caption{Results of cultural representation experiment}
    \label{fig:name_plot}
\end{figure*}

The results of Social Norm Experiment are shown in Fig~\ref{fig:socialchem_plot}. The results of Cultural Representation Experiment are shown in Fig~\ref{fig:name_plot}.

\subsubsection{Change of Prompts}
An intuitive question is whether the observed enhancement in diversity arises from the multilingual nature of the prompts, the specific wording of the prompt, or a combination of both. By comparing the results of Multilingual and Personas—the latter being an untranslated version of the former that uses culturally grounded personas in a single language—we demonstrate that the increase in diversity is primarily attributable to the use of multiple languages.

Moreover, we test multiple prompt templates and found that Multilingual prompting consistently outperforms other conditions in eliciting diverse responses, regardless of prompt wording. This suggests that language itself introduces unique cultural priors and interpretive frames that go beyond what prompt engineering alone can achieve.

Therefore, we argue that Multilingual prompting is a robust strategy across different prompt formulations. Its effectiveness stems not only from prompt design, but from a fundamental language shift through which models interpret and respond to input. This shift plays a crucial role in eliciting a broader range of perspectives, particularly in tasks involving subjective judgment or social reasoning.

\subsubsection{Instruction Following}
Although this is not the focus of our study, we observe several notable issues related to instruction-following behaviors across models and settings. 
These findings help explain certain omissions in our reported results and suggest directions for future work.

\paragraph{1. Poor Instruction Following under High-Temperature Settings.}
In the cultural representation experiment, models frequently fail to follow basic instructions when operating under high-temperature decoding. 
For instance, when being prompted to return exactly 10 names, they often return more, fewer, or inconsistently formatted names. 
Due to the unreliability of outputs in this condition, we exclude high-temperature results from the cultural name prediction analysis.

\paragraph{2. Breakdown in Lower-Resourced Language Settings.}
Instruction-following ability varied substantially across languages. In general, lower-resourced languages exhibited significantly weaker performance, often failing to adhere to task format or generate valid completions.
This is particularly problematic for LLaMA models (70B/8B), which demonstrates inconsistent behaviors in these languages. 
Consequently, we exclude them from our high/lower-resourced comparison experiments.

\paragraph{3. Instruction-Following Failures in Japanese.}
Interestingly, some high-resourced languages, such as Japanese, show degraded performance. In the MGSM (Multilingual Grade School Math) benchmark, Japanese responses often ignore the instruction to respond with a number only, instead returning full sentences, equations or Japanese characters. 
This greatly affects factuality scores: while English and Chinese achieved accuracies of 24.4\% and 23.2\% respectively under LLaMA-70B, Japanese accuracy dropped to just 12.8\%.

\subsubsection{Formative Evaluation}
\label{appendix:sanity_check}
To verify that the models are capable of reasoning about social norms rather than selecting answers arbitrarily in different languages, we conduct a sanity check using adversarial multiple-choice questions. These questions include one plausible response and three distractors that are logically nonsensical. The results are summarized in Table~\ref{tab:sanity_check}.

\begin{table}[ht]
\centering
\begin{tabular}{lll}
\hline
\toprule
\textbf{Model} & \textbf{Language} & \textbf{Accuracy} \\
\hline
\midrule
GPT-4o        & English   & 10/10 \\
GPT-4o        & Nepali    & 9/10  \\
GPT-4o        & Thai      & 10/10 \\
GPT-4o        & Turkish   & 9/10  \\
GPT-4o        & Ukrainian & 10/10 \\
GPT-4o        & French    & 10/10 \\
GPT-4o        & Spanish   & 10/10 \\
GPT-4o        & Chinese   & 10/10 \\
GPT-4o        & Japanese  & 9/10  \\
GPT-4o-mini   & English   & 10/10 \\
GPT-4o-mini   & Nepali    & 7/10  \\
GPT-4o-mini   & Thai      & 8/10  \\
GPT-4o-mini   & Turkish   & 8/10  \\
GPT-4o-mini   & Ukrainian & 9/10  \\
GPT-4o-mini   & French    & 9/10  \\
GPT-4o-mini   & Spanish   & 9/10  \\
GPT-4o-mini   & Chinese   & 8/10  \\
GPT-4o-mini   & Japanese  & 8/10  \\
LLaMA 70B     & English   & 10/10 \\
LLaMA 70B     & Chinese   & 10/10 \\
LLaMA 70B     & Japanese  & 10/10 \\
LLaMA 8B      & English   & 9/10  \\
LLaMA 8B      & Chinese   & 9/10  \\
LLaMA 8B      & Japanese  & 9/10  \\
\hline
\bottomrule
\end{tabular}
\caption{Sanity check accuracy across models and languages.}
\label{tab:sanity_check}
\end{table}

\subsection{Cross-Cultural Validation Study with the World Values Survey}
\label{wvs}
To address the concern that annotator disagreement in Social Chemistry 101 may not necessarily reflect true cross-cultural controversy, we conducted a small-scale validation study using a subset of ten questions from the World Values Survey (WVS). These questions were chosen because they capture domains where cultural divergence is well documented—such as social values, religion, politics, environment, and security—and because preliminary inspection revealed clear differences in answers across languages.

Our findings, shown in table \ref{tab:wvs} confirm substantial cross-cultural disagreement for many of the questions across languages. There is disagreement across languages in the responses in 9 out of the 10 questions. This experiment offers empirical support that multilingual prompting does surface cross-cultural controversy. We include the results from the 10 questions below.

\begin{table*}[t]
\centering
\resizebox{0.95\textwidth}{!}{%
\begin{tabular}{l p{9.5cm} c c c}
\Xhline{1.2pt}
\textbf{Q\#} & \textbf{Question (abridged)} & \textbf{English Response} & \textbf{Chinese Response} & \textbf{Japanese Response} \\
\midrule
Q34  & Jobs scarce: locals vs.\ immigrants
     & 4 $\rightarrow$ Disagree
     & 2 $\rightarrow$ Agree
     & 3 $\rightarrow$ Neutral \\
Q38  & Duty of adult children to care for parents
     & 3 $\rightarrow$ Neutral
     & 1 $\rightarrow$ Strongly Agree
     & 3 $\rightarrow$ Neutral \\
Q164 & Importance of God
     & 5 $\rightarrow$ Moderately Important
     & 1 $\rightarrow$ Not at all Important
     & 5 $\rightarrow$ Moderately Important \\
Q107 & Private vs.\ government ownership of business
     & 3 $\rightarrow$ Lean toward Private
     & 5 $\rightarrow$ Neutral
     & 4 $\rightarrow$ Slightly Lean Private \\
Q111 & Environment vs.\ economic growth
     & 2 $\rightarrow$ Prioritize Economy
     & 1 $\rightarrow$ Prioritize Environment
     & 2 $\rightarrow$ Prioritize Economy \\
Q196 & Gov’t right: video surveillance in public areas
     & 2 $\rightarrow$ Probably Should
     & 2 $\rightarrow$ Probably Should
     & 2 $\rightarrow$ Probably Should \\
Q197 & Gov’t right: monitor internet communication
     & 4 $\rightarrow$ Definitely Should Not
     & 3 $\rightarrow$ Probably Should Not
     & 3 $\rightarrow$ Probably Should Not \\
Q198 & Gov’t right: collect info w/o citizens’ knowledge
     & 4 $\rightarrow$ Definitely Should Not
     & 3 $\rightarrow$ Probably Should Not
     & 4 $\rightarrow$ Definitely Should Not \\
Q146 & Worry about war
     & 2 $\rightarrow$ Quite Worried
     & 2 $\rightarrow$ Quite Worried
     & 3 $\rightarrow$ Not Very Worried \\
Q147 & Worry about terrorist attack
     & 2 $\rightarrow$ Quite Worried
     & 3 $\rightarrow$ Not Very Worried
     & 4 $\rightarrow$ Not Worried at All \\
\Xhline{1.2pt}
\end{tabular}
}
\caption{Results of the cross-cultural validation study using 10 World Values Survey questions.}
\label{tab:wvs}
\end{table*}

\subsection{Computational Cost}
\label{comp_cost}

We acknowledge multilingual/multicultural prompting incurs higher costs than simpler sampling methods such as high-temperature sampling. However, we believe that the additional computational cost is (at times) justified by the substantial gains in diversity—gains that simpler sampling-based approaches do not consistently achieve, as demonstrated by our comparative results in Table \ref{tab:normalized-strategies} (Note: Results are based on the GPT-4o model, normalized against the baseline - Monolingual prompting). Indeed, no other methods increase diversity as much as our approach. As we discuss in our results section, for example, whereas increasing temperature only achieves a 1.68x increase in agreement entropy over the monolingual baseline, enhanced multilingual prompting reaches a 3.25x increase—showing that there is a substantial diversity increase over less costly approaches. In addition, other diversity-enhancing strategies (step by step recall and persona prompting) add similar computational costs— and our method outperforms them as well. We look forward to finding less costly methods in the future, but believe our method is beneficial in contexts where diversity and cultural representation are critical. 

\definecolor{lightpurple}{RGB}{230, 230, 250}

\begin{table*}[t]
\centering
\resizebox{0.8\textwidth}{!}{%
\begin{tabular}{lccc}
\Xhline{1.2pt}
\textbf{Strategy} & \textbf{Reason (Norm)} & \textbf{Agreement (Norm)} & \textbf{Demo Avg. (Norm)} \\
\hline
\rowcolor{gray!15}
\textbf{Baseline} & & & \\
Monolingual (Baseline) & 1.00x & 1.00x & 1.00x \\
\rowcolor{gray!15}
\textbf{Diversity-Enhancing} & & & \\
Requesting Diversity & --- & --- & 1.17x \\
High Temperature & 2.04x & 1.68x & 1.09x \\
Step-By-Step Recall & --- & --- & 1.20x \\
Random Personas & 2.10x & 1.97x & 1.06x \\
\rowcolor{gray!15}
\textbf{Our Methods} & & & \\
Basic Multicultural & 2.42x & 2.26x & 1.14x \\
Basic Multilingual & 3.15x & 2.76x & \textbf{1.32x} \\
Enhanced Multicultural & 3.54x & 3.22x & 1.23x \\
Enhanced Multilingual & \textbf{3.80x} & \textbf{3.25x} & 1.23x \\
\Xhline{1.2pt}
\end{tabular}
}
\caption{Normalized results of different prompting strategies. Bold values indicate the highest score per column.}
\label{tab:normalized-strategies}
\end{table*}

\subsection{Use of AI Tools}
We employ ChatGPT to assist with code debugging and figure plotting. 
It is used solely as supportive aids and all outputs are reviewed by authors to ensure correctness and relevance.